\documentclass[lettersize,journal]{IEEEtran}
\usepackage{amsmath,amsfonts}
\usepackage{amsthm}
\usepackage{algorithmic}
\usepackage{algorithm}
\usepackage{array}
\usepackage[caption=false,font=normalsize,labelfont=sf,textfont=sf]{subfig}
\usepackage{textcomp}
\usepackage{stfloats}
\usepackage{url}
\usepackage{verbatim}
\usepackage{graphicx}
\usepackage{cite}
\usepackage{xfrac}
\usepackage{bm}
\usepackage{threeparttable}
\usepackage{tabu}
\usepackage{multirow}
\usepackage{graphicx}
\usepackage{bbding}
\usepackage{color}
\usepackage{pifont}

\usepackage[hidelinks, colorlinks=true,linkcolor=blue]{hyperref}

\newcounter{magicrownumbers}
\newcommand\rownumber{\stepcounter{magicrownumbers}\arabic{magicrownumbers}}

\newtheorem{theorem}{Theorem}

\newtheorem{lemma}[theorem]{Lemma}

\begin{document}

\title{SoftSignSGD(S3): An Enhanced Optimizer for Practical DNN Training and Loss Spikes Minimization Beyond Adam}

\author{Hanyang Peng, ~\IEEEmembership{Member,~IEEE,}
        Shuang Qin,
        Yue Yu,~\IEEEmembership{Member,~IEEE,}
        Fangqing Jiang,
        Hui Wang,
        and Wen Gao, ~\IEEEmembership{Fellow,~IEEE}
        % <-this % stops a space
\thanks{Hanyang Peng, Shuang Qin, Yue Yu, Fangqing Jiang, and Hui Wang are with Pengcheng Laboratory, Shenzhen 518066, China (e-mail: philoso\_phy0922@163.com; qinsh@pcl.ac.cn; yuy@pcl.ac.cn;  jiangfq@pcl.ac.cn;  wangh06@pcl.ac.cn).

Wen Gao is with the Pengcheng Laboratory, Shenzhen 518066, China, and also with Peking University, Beijing 100871, China,(e-mail: gaow@pcl.ac.cn).

Corresponding author: Yue Yu.
}% <-this % stops a space
%\thanks{Manuscript received April 19, 2021; revised August 16, 2021.}
}

\markboth{Journal of \LaTeX\ Class Files, ~2025}%
{Shell \MakeLowercase{\textit{et al.}}: A Sample Article Using IEEEtran.cls for IEEE Journals}

%\ieeepubid{0000--0000/00\$00.00~\copyright~2021 ieee}

\maketitle

\begin{abstract}
\textsf{\small Adam} has proven remarkable successful in training deep neural networks, but the mechanisms underlying its empirical successes and limitations remain underexplored. In this study, we demonstrate that the effectiveness of \textsf{\small Adam} stems largely from its similarity to \textsf{\small SignSGD} in robustly handling large gradient fluctuations, yet it is also vulnerable to destabilizing loss spikes due to its uncontrolled update scaling. To enhance the advantage of Adam and mitigate its limitation, we  propose \textsf{SignSoftSGD} (\textsf{S3}), a novel optimizer with three key innovations. \emph{First}, \textsf{S3} generalizes the sign-like update by employing a flexible $p$-th order momentum ($p \geq 1$) in the denominator, departing from the conventional second-order momentum (variance) preconditioning. This design enables enhanced performance while achieving stable training even with aggressive learning rates.  \emph{Second}, \textsf{S3} minimizes the occurrences of  loss spikes through unified exponential moving average coefficients for numerator and denominator momenta, which inherently bound updates to $[-1, 1]$ and simplify hyperparameter tuning. \emph{Third}, \textsf{S3} incorporates an equivalent  Nesterov's accelerated gradient(NAG) module, accelerating convergence without memory overhead.  Theoretically, we prove that \textsf{S3} achieves the optimal convergence rate of $O\left(\frac{1}{T^{\sfrac{1}{4}}}\right)$ for general nonconvex stochastic optimization under weak assumptions. Extensive experiments across a range of vision and language tasks show that \textsf{\small S3} not only converges more rapidly and improves performance but also rarely experiences loss spikes, even with a \textbf{$\bm{10 \times}$} larger learning rate. In fact, \textsf{\small S3} delivers performance comparable to or better than \textsf{\small AdamW} with \textbf{$2 \times$} the training steps, establishing its efficacy in both efficiency and final task performance.
\end{abstract}

\begin{IEEEkeywords}
Optimizer, Loss spike, Adam
\end{IEEEkeywords}

\section{Introduction}
\IEEEPARstart{O}ptimizers are critical components in the training of deep neural networks (DNNs). Among them, \textsf{\small Adam}\cite{Adam2015} has emerged as the predominant choice for training Transformers~\cite{Transformer_2017}, particularly for state-of-the-art large language models (LLMs) such as GPT-3~\cite{GPT32020}, PaLM~\cite{Palm2023}, and LLaMA~\cite{Llama3_2024}, DeepSeek ~\cite{DeepSeek_V3_2024},  as well as large vision models like CLIP~\cite{CLIP_2021} and SAM~\cite{SAM_2023}. Notably, Adam has also become the de facto optimizer for modern convolutional neural networks (CNNs), including ConvNeXt~\cite{ConvNext_2022,ConvNext2_2023}, despite the historical preference for stochastic gradient descent (\textsf{\small SGD}) in classic-CNN training~\cite{AlexNet2013,ResNet2016}.

\begin{figure*}[t]
  \centering
  \hspace{-8pt}
  \subfloat[\footnotesize{Trajectories in 3D space}]{\includegraphics[width=.26\textwidth]{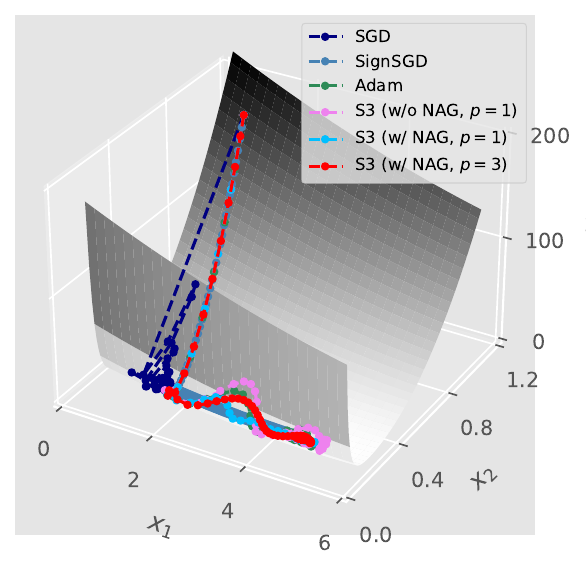}} \hspace{5pt}
  \subfloat[\footnotesize{Dynamics of Loss Convergence}]{\includegraphics[width=.35\textwidth]{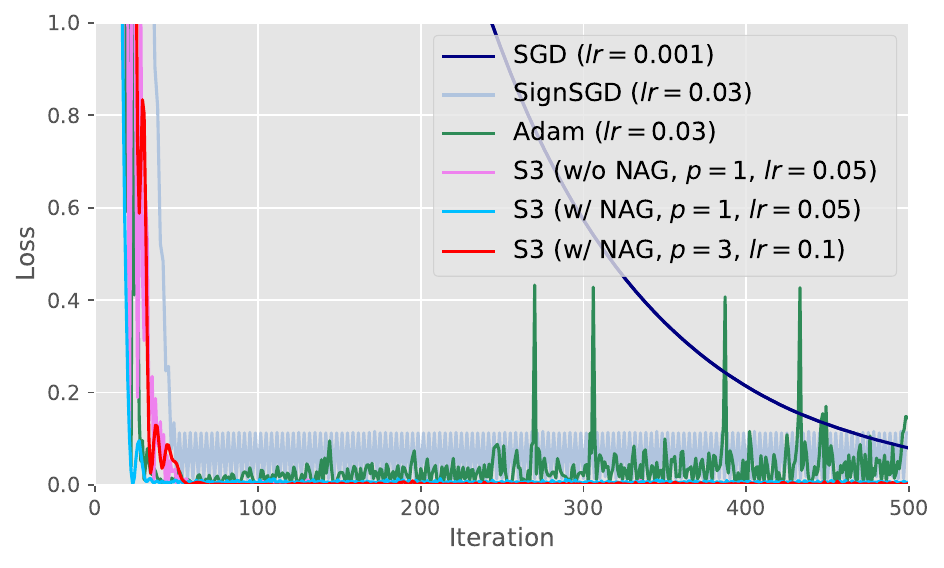}} \hspace{5pt}
  \subfloat[\footnotesize{Dynamics of Mean update} ]{\includegraphics[width=.35\textwidth]{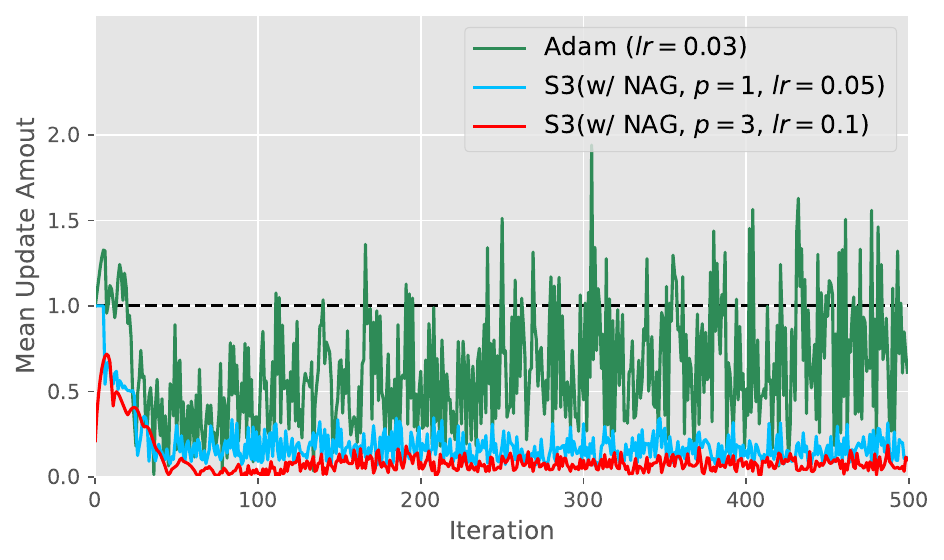}} \\
  \caption{{ The (a) trajectories, (b) loss convergence curves, and (c) mean update curves of \textsf{\footnotesize SGD}, \textsf{\footnotesize SignSGD}, \textsf{\footnotesize Adam} and \textsf{\footnotesize S3}. The loss function is defined as $f(x^{(1)}, x^{(2)})=0.5(x^{(1)} - 1/x^{(2)})^2 + 0.5(x^{(1)} - 20x^{(2)})^2 $. The initial point is set as $[x_0^{(1)}, x_0^{(2)} ] = [1.0, 1.0]$. Gaussian noise is added to the gradient at each iteration to simulate random sampling, represented as $g_t = [\nabla {\footnotesize f_{x_t^{(1)}}}, \nabla f_{ x_t^{(2)}}] + {\mathcal N}(0, 0.1)$.
  Due to the significant gap between $\nabla f_{x_t^{(1)}}$ and $\nabla f_{x_t^{(2)}}$, we set a small learning rate  for \textsf{\footnotesize SGD} to prevent divergence. However, this results in slow convergence. The update of \textsf{\footnotesize SignSGD}, ${\rm Sign}(g_t)$, causes the loss to oscillate and prevents it from converging to the minimum.
  The sign-like property of \textsf{\footnotesize Adam} makes it perform much better than \textsf{\footnotesize SGD}. The update of \textsf{\footnotesize Adam}, $\frac{m_t}{\sqrt{v_t}}$, is generally smaller, aiding in achieving a lower loss, compared to \textsf{\footnotesize SignSGD}. However, it has a non-trivial probability of encountering loss spikes.
  The update of \textsf{\footnotesize S3} is constrained to $[-1,1]$ and gradually diminishes as the loss approaches the minimum. This property enables \textsf{\footnotesize S3} to achieve an extremely small loss and seldom encounter loss spikes.
  The introduction of the NAG technique in \textsf{\footnotesize S3} is helpful for accelerating convergence. The use of a large $p$-th order momentum allows \textsf{\footnotesize S3} to employ a large learning rate  without encountering training instability.    }}
  \label{Fig.1}
\end{figure*}

%This raises two open questions:
%
%
%\vspace{0pt}
%
%\emph{What are the essential factors contributing to the practical success of { Adam}? And is there room for further improvement in vanilla { Adam}?}
%
%\vspace{0pt}

%The updating rule of \textsf{\small Adam} is defined as $\bm{x}_{t+1} =  bm{x}_t - \gamma_t \frac{\bm{m}_{t}}{\sqrt{\bm{v}_t}}$ where $\bm{m}_{t} = \beta_1 \bm{m}_{t-1} + (1-\beta_1)\bm{g}_t$ and $\bm{v}_{t} = \beta_2 \bm{v}_{t-1} + (1-\beta_2)\bm{g}_t^2$,

While the practical success of \textsf{\small Adam}~\cite{Adam2015} is indisputable in deep learning, the theoretical underpinnings of its effectiveness remain inadequately understood. A prevailing hypothesis attributes its performance to second-order momentum (variance) preconditioned adaptivity originating from the original paper on Adam ~\cite{Adam2015}. This adaptivity is exemplified by the update rule: $\bm{x}_{t+1} = \bm{x}_t - \frac{\gamma_t}{\sqrt{\bm{v}_t}} \circ \bm{m}_t$
which employs the second-order momentum  $\sqrt{\bm{v}_t}$ to adaptively precondition $\bm{m}_t$.  However, recent work by~\cite{Lion2023} challenges this long-held assumption. Their proposed optimizer, \textsf{\small Lion}, demonstrates that eliminating the second-moment preconditioning entirely can yield comparable---and occasionally superior---performance to Adam across diverse DNN architectures and tasks. This raises questions about whether Adam's efficacy stems from factors beyond its canonical design principles.

To investigate \textsf{\small Adam}'s success, we analyze its update dynamics. Notably, the term $\frac{\bm{m}_{t}}{\sqrt{\bm{v}_t}}$ in \textsf{\small Adam}'s update exhibits \emph{quasi-binary behavior}, approximating a sign-descent direction. Empirical studies~\cite{Explian_Adam_SGD_2023, Explian_Adam_SGD_2024} show that explicit sign-based updates can nearly close the performance gap between \textsf{ \small SGD} and \textsf{\small Adam} in training complex models like Transformers, suggesting \textsf{\small Adam}'s implicit sign-like behavior is pivotal to its success. However, prior work leaves unresolved why sign-descent proves particularly effective for Transformers. Our analysis identifies \emph{inter- and intra-layer gradient heterogeneity} as a critical factor: Transformers exhibit pronounced disparities in gradient magnitudes across layers and parameters, which Adam's conservative, sign-like updates inherently mitigate.
On the other hand, we also reveal limitations in \textsf{\small Adam}'s design. Specifically, Adam's update magnitudes can grow disproportionately large with non-negligible probability, triggering destabilizing chain reactions that correlate with training instability and loss spikes. This phenomenon underscores a trade-off between \textsf{\small Adam}'s robustness to gradient heterogeneity and its susceptibility to unstable updates.

Building on insights gained from our analysis of \textsf{\small Adam}, we introduce a novel optimizer, \textsf{\small SignSoftSGD (S3)}.\emph{First}, \textsf{S3} generalizes the sign-like update mechanism by replacing the second-order conventional momentum denominator with a $p$-th order momentum ($p \geq 1$). This break from the routine enables \textsf{\small S3} to leverage a larger $p$-order momentum, facilitating faster convergence and improved performance with larger learning rates, while avoiding the training instability associated with \textsf{\small Adam}. \emph{Second}, \textsf{\small S3} uses the same exponential moving average coefficient $\beta$ for both the numerator and denominator momenta in the update, strictly constraining the update to the range $[-1, 1]$\footnote{This is the reason we named the optimizer \textsf{\small SoftSignSGD}.}. This design minimizes the occurrence of loss spikes by limiting the maximum update. Additionally, \textsf{\small S3} offers practical advantages, including the elimination of bias correction and gradient clipping, and reduces tuning complexity by removing one hyperparameter. \emph{Third}, \textsf{\small S3} integrates an equivalent of Nesterov's accelerated gradient (NAG), further enhancing training speed without incurring additional memory costs. We illustrate the convergence behaviors of \textsf{\small SGD}, \textsf{\small SignSGD}, \textsf{\small Adam}, and \textsf{\small S3} in Figure~\ref{Fig.1}. Furthermore, we provide a theoretical analysis of \textsf{\small S3}'s convergence rate for a general nonconvex stochastic optimization problem, demonstrating that it achieves the optimal lower bound on the convergence rate of $O\left(\frac{1}{T^{\sfrac{1}{4}}}\right)$, where $T$ denotes the number of iterations, under weak assumptions.

Our primary contributions are summarized as follows:
\begin{itemize}
  \item {We theoretically and empirically demonstrate that \textsf{\small Adam} is the primary cause of loss spikes when training large models (\emph{i.e.}, LLMs)}, due to its potential for excessively large parameter updates.
  \item {We introduce a novel optimizer, \textsf{\small S3}, which offers four key advantages over \textsf{\small Adam}}:
    \begin{enumerate}[]
      \item Elimination of bias correction and gradient clipping, reducing one hyperparameter;
      \item A generalized sign-like formulation enabling larger learning rates for improved performance;
      \item Integration of an equivalent NAG technique to speed up training convergence without additional memory overhead;
      \item Minimization of the risk of training instability and loss spikes.
    \end{enumerate}
  \item We provide a theoretical analysis of \textsf{\small S3} on a general nonconvex stochastic problem, {achieving the optimal convergence rate under a weak non-uniform smoothness assumption.}
  \item We conduct extensive experiments to compare \textsf{\small S3} with \textsf{\small Adam} and other optimizers. The results show that \textsf{\small S3} offers faster training and superior performance, {achieving results comparable to or better than \textsf{\small AdamW} with twice the training steps, while rarely experiencing loss spikes, even at significantly higher learning rates.}
\end{itemize}

\section{Related Work}

\textbf{ Optimizers in Deep Learning.} Nowadays, \textsf{\small Adam} has become the dominant optimizer in deep learning.
The adaptivity strategy in \textsf{\small Adam} traces its roots back to earlier optimizers such as \textsf{\small Adagrad} \cite{AdaGrad2011}, \textsf{\small RMSprop} \cite{RMSProp2012}, and \textsf{\small Adadelta} \cite{Adadelta2012}. Beyond \textsf{\small Adam}, a wide range of  variants  are proposed \cite{NAdam2016,AMSGrad2018,AdamW2017,AdaBelief2020,Adafactor2018}. \textsf{\small SignSGD}, the first sign descent method, was proposed to reduce communication costs in distributed learning \cite{1-bitSGD2014}. Subsequently, \cite{SignSGD2018, SignSGD_2023} provided theoretical convergence for \textsf{\small SignSGD} and introduced an enhanced version.  \cite{Lion2023} applied an auto ML method to discover the sign descent optimizer \textsf{\small Lion}.  This optimizer demonstrated improved performance with a faster convergence rate on various tasks compared to \textsf{\small Adam}. Recently, \cite{Sophia2023} introduced an effective second-order optimizer for LLM pre-training.

\textbf{Sign-based Optimizers.} Sign-based algorithms that simply exploits the signs of gradients could date back to \textsf {\small RPROP} \cite{RPROP_1993}. \cite{1-bit-SGD_2014} proposed \textsf {\small 1-bit SGD} and empirically demonstrate it achieve good performance while dramatically reducing the communication costs in distributed system.  \cite{Lion2023} employs an AutoML method to discover an effective optimizer, \textsf {\small Lion}, resembling  \textsf{\small SignSGD} with momentum, and demonstrate superior performance to Adam across diverse DNN models. \cite{Lion_explain_2023}  theoretically analyzed the efficacy of \textsf {\small Lion} from the perspective of weight decay.  Meanwhile, the original version of \textsf {\small Adam }, \textsf {\small RMSProp} \cite{RMSProp2012}, were developed from the sign-based Rprop.  \cite{MSSD_2018} also found that sign descent algorithms has a deep connection with \textsf {\small Adam}. Recently, \cite{Explian_Adam_SGD_2023,Explian_Adam_SGD_2024} empirically showcase that the sign-like property of \textsf {\small Adam} is just the primary reason behind its superior performance for training DNNs.

\textbf{Nesterov's Accelerated Gradient (NAG).} Theoretical demonstrations by \cite{NAG1983, NAG2013} indicate that NAG can achieve faster convergence on convex optimization problems compared to vanilla gradient descent, leveraging gradient information at an extrapolation point to anticipate future trends.
 \textsf{\small NAdam} by \cite{NAdam2016} was the first to incorporate NAG into adaptive optimizers, modifying the first-order momentum of \textsf{\small Adam} with NAG.  \textsf{\small Adan} by \cite{Adan2024} integrated a equivalence of NAG into both the first and second momentum of \textsf{\small Adam}, and \textsf{\small Win} \cite{Win_2023} applied Nesterov acceleration to the update rather than the first and second momentum.  \textsf{\small Adan} and \textsf{\small Win} outperformed \textsf{\small Adam} on various tasks, but they require tuning additional hyperparameters and consume more memory, compared to vanilla \textsf{\small Adam}. \textsf{\small Lion} \cite{Lion2023}, despite being a sign descent method, exhibits a momentum construction similar to NAG \cite{Lion_explain_2023}. This resemblance could be a contributing factor to its superior speed and performance over \textsf{\small Adam}.

\textbf{Training instability and Loss Spikes in LLM Training.} Encountering loss spikes is a common phenomenon during LLM training \cite{GLM130B2022,Palm2023,Llama2023,Baichuan22023}. However, the underlying reasons for this problem were not well explored prior to this. Practitioners had to resort to ad hoc engineering strategies such as skipping some data batches before the spike occurs and restarting training from a nearby checkpoint \cite{Palm2023,Adaminstability2023}, resulting in resource wastage due to frequent rollbacks and checkpointing savings. Some previous works  investigated the phenomenon of train instability \cite{RAdam_2019} and loss spikes \cite{Catapults_SGD_2023, Loss_spike_2023}. \cite{RAdam_2019}  demonstrated that the variance of the update $\sfrac{1}{\sqrt{\bm{v}_t^{(j)}}}$ is significantly larger, often causing the update to become disproportionately large, but the analysis only works in the early stage. The analyses in \cite{Catapults_SGD_2023, Loss_spike_2023}  were restricted to either linear models or shallow networks with mean squared error (MSE) loss, using (S)GD as the optimizer. Consequently, it is questionable whether these findings can be directly applied to the context of LLM training.  Previous attempts to mitigate instability include embedding norm with BF16, but this comes at a significant performance tax \cite{scaolanguage2022}. Some researchers found that gradient shrink on the embedding layer reduces loss spikes \cite{GLM130B2022}. Others suggest normalizing the output embedding to lower spike risks \cite{Baichuan22023}.  More recently, \cite{Adaminstability2023} suggested that  time-domain correlation between gradient estimates of earlier layers contributes to loss spikes during LLM training. The suggested mitigation strategies include lowering the $\epsilon$ value in \textsf{\small Adam} and reducing the batch size. However,  the study itself that these methods are not silver bullets for a fundamental solution.

%\textbf{To the best of our knowledge, the analyses presented above provide the first formal explanation for the frequent occurrence of loss spikes during LLM training.}

\section {Rethinking the Effectiveness and Ineffectiveness of Adam}

In a deep learning task, the optimizer aims to minimize the empirical risk loss of a model on a dataset, \emph{i.e.},
{\small
\begin{equation}
\min_{x\in {\mathbb R}^d } F({\bm{x}}) = {\mathbb E}_{\zeta\sim {\mathcal D}}[f(\bm{x};\zeta)] =  \frac{1}{n}\sum_{i=1}^n f(\bm{x};\omega_i),
\label{Eq.obj}
\end{equation}
} where $\bm{x}$ is the $d$-dimensional model parameter, and $\zeta$ is independently and identically sampled from the dataset $\{\omega_i: \omega_i \in {\mathcal D}, 1\le i \le n \}$.

\textsf{\small Adam} has become the dominant optimizer for training deep neural networks (DNNs), significantly outperforming \textsf{\small SGD} in training the increasingly popular Transformer models. Its remarkable efficacy extends even to CNN-based models like ConvNeXT, where \textsf{\small Adam} is preferred over \textsf{\small SGD} for achieving superior performance, despite the historical belief that \textsf{\small SGD} is better suited for training CNNs. While the practical success of \textsf{\small Adam} is indisputable, the underlying factors driving its effectiveness remain largely unexplored. Understanding these factors is crucial for making further advancements in DNN training.

%The main reason why \textsf{Adam} can effectively train complex DNN models have been actually scattered in the previous research works. We gather these works up to unveil the reason in this section.

Recalling the updating rule of \textsf{\small Adam}, we have
{\small
\begin{equation}
\begin{aligned}
&\tilde{\bm{m}}_{t} = \beta_1 \tilde{\bm{m}}_{t-1} + (1-\beta_1)\bm{g}_t, \\
&\bm{m}_{t} = \frac{\tilde{\bm{m}}_t}{1-\beta_1^t}, \\
&\tilde{\bm{v}}_{t} = \beta_2 \tilde{\bm{v}}_{t-1} + (1-\beta_2)\bm{g}_t^2, \\
&\bm{v}_{t} = \frac{\tilde{\bm{v}}_t}{1-\beta_2^t}, \\
&\bm{x}_{t+1} =  \bm{x}_t - \gamma_t \frac{\bm{m}_{t}}{\sqrt{\bm{v}_t}},
\end{aligned}
\label{Eq.Adam}
\end{equation}
}where $x_t$ denotes the model parameter, $\bm{g}_t= \nabla f(\bm{x}_{t};\zeta_t)$ is the stochastic gradient, $\gamma_t$ is the learning rate, and $\beta_1$ and $\beta_2$ represents the exponential moving average coefficients.

In essence, $\vert \bm{m}_t \vert$ and $\sqrt{\bm{v}_t}$ are of the same order of magnitude. Specifically,  if $g_t$ ideally stays stable over a period, \textsf{\small Adam} in Eq. (\ref{Eq.Adam}) reduces  to \textsf{\small SignSGD}, \emph{i.e.}. $\bm{x}_{t+1} = \bm{x}_t - \gamma_t \frac{\bm{m}_{t}}{\sqrt{\bm{v}_t}} =  \bm{x}_t - \gamma_t {\rm Sign}(\bm{g}_t) $. Therefore, \textsf{\small Adam} can be viewed as a sign-like optimizer.

The primary reason behind \textsf{\small Adam}'s practical effectiveness over \textsf{\small SGD} in training complex DNNs remains fragmented across prior studies and lacks comprehensive consolidation. \cite{Explian_Adam_SGD_2023} empirically demonstrates that sign descent with momentum yields performance comparable to \textsf{\small Adam} when training Transformers, though it lacks a thorough analytical justification. More recently, \cite{Lion2023} presents an AutoML approach to discover a new optimizer, \textsf{\small Lion}, which resembles \textsf{\small SignSGD} with momentum and outperforms \textsf{\small Adam} across various DNN models. Both \textsf{\small Adam} and \textsf{\small Lion} owe their effectiveness primarily to their shared sign-like property.

For deep networks, gradients of the initial and final layers often differ significantly, as theoretically verified in \cite{BN_mean_field_2019, Understanding_difficulty_2020, LN_mean_field_2020, lipschitz_2021, Lipsformer_2023} through mean-field theory and Lipschitz continuity. Additionally, within the same Transformer layer, gradients can vary considerably due to the attention mechanism \cite{Signal_propa_2022}. This substantial gradient discrepancy challenges \textsf{\small SGD}, which directly uses gradients for updates. To avoid divergence, \textsf{\small SGD} requires a small learning rate, resulting in slower training. Another drawback is that parameters with large gradients undergo substantial updates, while those with small gradients remain nearly unchanged, weakening the network's overall representation capacity and degrading final performance.

In contrast, \textsf{\small Adam} maintains updates close to $\pm1$ despite significant gradient disparities, thanks to its inherent sign-like property. This conservative update behavior makes \textsf{\small Adam} an effective optimizer for training complex DNNs. \textbf{In summary, the efficacy of \textsf{\small Adam} in training complex DNNs arises from its conservative sign-like descent, which mitigates the impact of significant gradient discrepancies.}

 \begin{figure}[h]
 \centering
  \subfloat{\includegraphics[width=.36\textwidth]{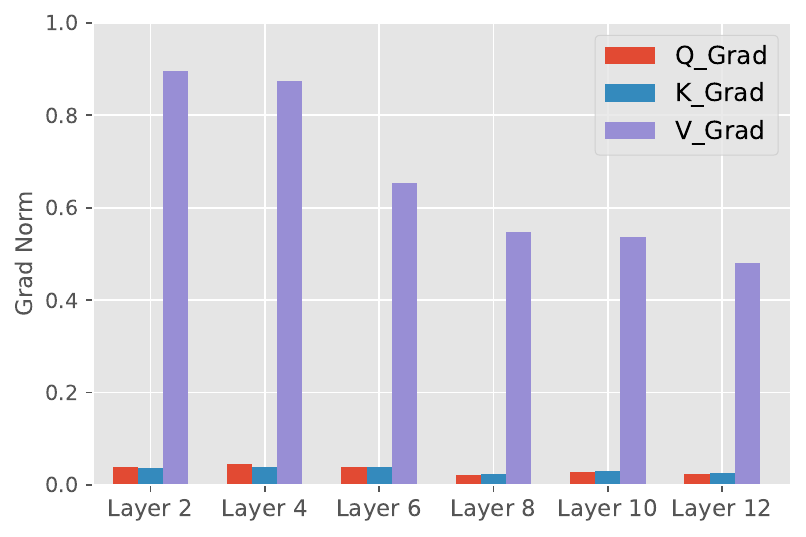}}
  \caption{\small{ Visualization of gradient norms within different layers in ViT-B/16 at initialization.  } }
  \label{Fig.2}
\end{figure}

%While the performance of \textsf{\small SignSGD} closely approximates \textsf{\small Adam} in training DNNs, it remains inferior, indicating that \textsf{\small Adam} possesses additional mechanisms contributing to its effectiveness beyond the sign-like property.   When the  parameter $x_t$ approaches a local optimum, an ideal optimizer should ensure that updates gradually decay to zero. Failure to do so may result in $x_t$ oscillating around the optimum without effectively converging to it.  However, the update of \textsf{\small SignSGD}, \emph{i.e.}, ${\rm Sign}(g_t)$, keeps $\pm1$ throughout training.  Consequently, the parameter $x_{t}$ bounds around the minimum. Conversely, the updating rule of \textsf{\small Adam}, as illustrated in Eq. ({\ref{Eq.Adam}}), exhibits a distinct behavior. As the elements of $g_t$ change signs more frequently in the latest periods before $t$, $\vert m_t \vert$ commonly becomes smaller, while $v_t$ remains unaffected.  In the final training stage, the elements of $g_t$ frequently change signs in the neighborhood of the optimal gradient $g^*=0$. This behavior implies that the update of \textsf{\small Adam}, denoted as $\frac{m_t}{\sqrt{v_t}}$, will likely be close to zero with high probability, facilitating the parameter's approach to the locally minimal point. In conclusion, \textsf{\small Adam} is more likely to aid loss convergence to a smaller value and achieve superior performance, compared with \textsf{\small SignSGD}.

While \textsf{\small Adam} effectively trains complex DNNs, it also escalates the risk of training instability and loss spikes with non-trivial probability.  This can be inferred from Theorem \ref{them.1}.

\begin{theorem}
The sequences $\{\bm{m}_t\}$ and $ \{\bm{v}_t\}$ are generated by Adam in Eq. (\ref{Eq.Adam}). If the moving average coefficients satisfy $\beta_1^2 <
\beta_2$,   then it holds that
\begin{equation}
\frac{\vert\bm{m}_t^{(j}\vert}{\sqrt{\bm{v}_t^{(j)}}} \le \frac{(1-\beta_1)\sqrt{1-\beta_2^t}\sqrt{1-(\frac{\beta_1^2}{\beta_2})^t}}{(1-\beta_1^t)\sqrt{1-\beta_2}\sqrt{1-\frac{\beta_1^2}{\beta_2}}} \simeq \frac{1-\beta_1}{\sqrt{1-\beta_2}\sqrt{1-\frac{\beta_1^2}{\beta_2}}},
\end{equation}
where $\frac{\vert\bm{m}_t^{(j}\vert}{\sqrt{\bm{v}_t^{(j)}}}$  reach to the largest value if the signs of $\{\bm{g}_t^{(j)}, \bm{g}_{t-1}^{(j)}, ...\bm{g}_{t-k}^{(j)}...\}$ are the same and $\vert\bm{g}_t^{(j)}\vert = \frac{\beta_2\vert\bm{g}_{t-1}^{(j)}\vert}{\beta_1} = \frac{\beta_2^2\vert\bm{g}_{t-2}^{(j)}\vert}{\beta_1^2} = ...=\frac{\beta_2^k\vert\bm{g}_{t-k}^{(j)}\vert}{\beta_1^k}...$\footnote{The update $\sfrac{\bm{m}_t^{(j)}}{\sqrt{\bm{v}_t^{(j)}}}$ with respect to $\{\bm{g}_k^{(j)}\}_{k=1}^t$ is a continuous function. Thus, when most of the signs of $\{\bm{g}_k^{(j)}\}_{k=1}^t$ are consistent and the secondary condition is nearly satisfied, $\sfrac{\bm{m}_t^{(j)}}{\sqrt{\bm{v}_t^{(j)}}}$ will be close to the theoretical maximum.}.
\label{them.1}
\end{theorem}

 \begin{figure*}[!th]
 \centering
  \subfloat{\includegraphics[width=.7\textwidth]{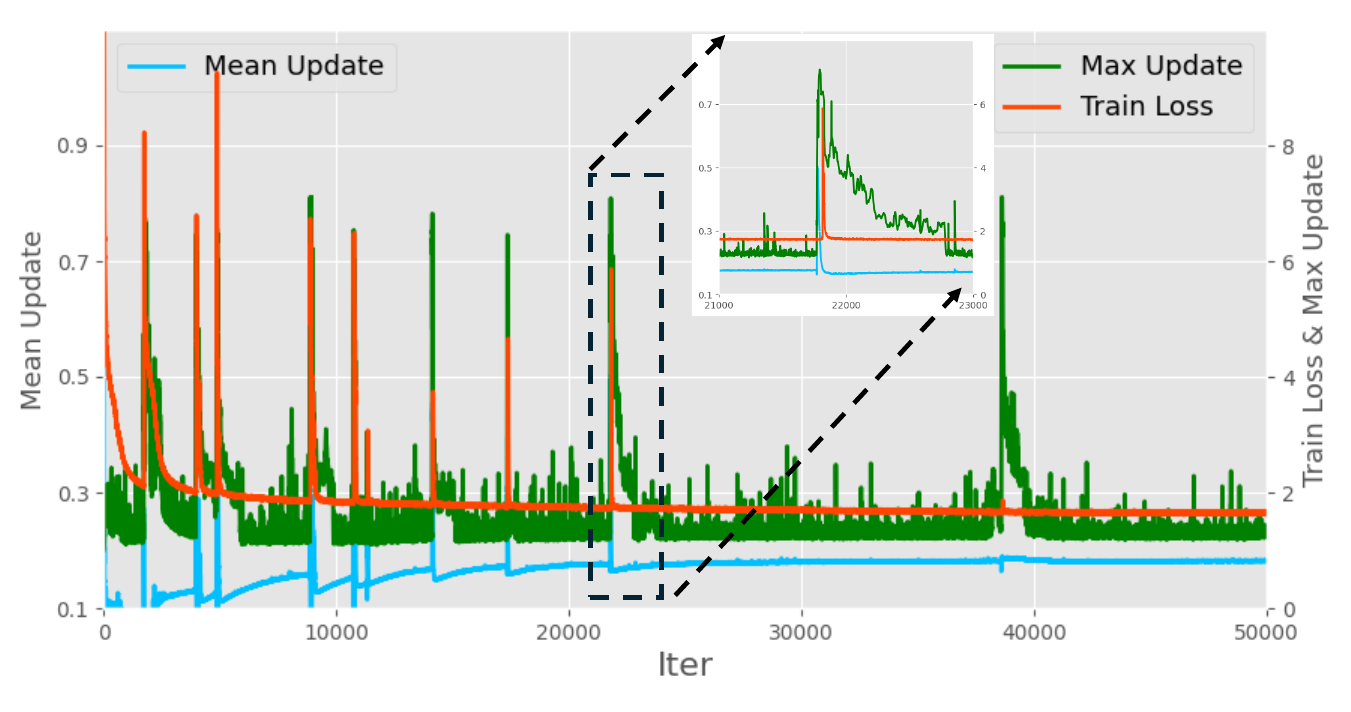}}
  \caption{\small{
Visualization of the mean update (\emph{i.e.}, $\mathrm{Avg}(\sfrac{\vert \bm{m}_t^{(j)} \vert}{\sqrt{\bm{v}_t^{(j)}}})$), the maximum update (\emph{i.e.}, $\max_{j \in [d]} (\sfrac{\vert \bm{m}_t^{(j)} \vert}{\sqrt{\bm{v}_t^{(j)}}})$), and the training loss over 50,000 iterations during GPT-2 (345M) training on OpenWebText using AdamW ($\beta_1 = 0.9, \beta_2 = 0.999$) with a cosine learning rate schedule. The figure illustrates that all loss spikes are preceded by abrupt increases in the mean update, following a sharp rise in the maximum update. This suggests that a sudden increase in the maximum update for any coordinate can lead to a significant rise in the mean update, which then triggers loss spikes. Moreover, these spikes primarily occur during the early training phase when the learning rate is relatively high. In later stages, as the learning rate decreases, loss spikes become infrequent, even with large maximum updates, like around Iteration 40,000.} }
  \label{Fig.2_1}
\end{figure*}

Theorem \ref{them.1} shows that when \textsf{\small Adam} is used, there is a probability that the update for each element $\frac{\vert\bm{m}_t^{(j)}\vert}{\sqrt{\bm{v}_t^{(j)}}}$ can become excessively large. For example, with typical values of $\beta_1=0.9$ and $\beta_2=0.999$, this update can approach its theoretical maximum of $\frac{1-\beta_1}{\sqrt{1-\beta_2}\sqrt{1-\frac{\beta_1^2}{\beta_2}}}\simeq 7.27$, while the normal absolute value of the update is typically much smaller than 1. Though the probability of any specific parameter's update reaching this maximal value is low, the likelihood that at least one parameter's update exceeds this threshold is high, particularly in large models (\emph{e.g.}, LLMs), which contain millions or even billions of parameters. When a parameter's update is excessively large and the learning rate is also high, it is likely to deviate substantially from its intended trajectory. Such deviations can propagate through the model's interconnected parameters, triggering a chain reaction that results in loss spikes. This mechanism helps explain the frequent loss spikes observed during the training of LLMs, particularly in the early stages when learning rates are higher. Specifically, the probability of loss spikes increases with the size of the LLM.

\textbf{In conclusion, vanilla \textsf{\small Adam} poses a significant risk of loss spikes during large-scale model training. Mitigating this problem requires strategies to constrain the maximum update magnitude for each parameter coordinate.}

\section { S3 Algorithm}

Analyzing \textsf{\small Adam} provides valuable insights that inform the design of a more effective optimizer for DNNs. \emph{First}, the update rule of the optimizer need only approximate the sign of the gradient, without the necessity of strictly following the formulation involving the ratio of first-order gradient momentum to second-order gradient momentum. \emph{Second}, minimizing the largest possible value of the update is crucial for mitigating potential loss spikes.

Furthermore, recent studies \cite{NAdam2016,Adan2024,Win_2023} have incorporated the NAG technique into DNN optimizers, consistently demonstrating faster training convergence and improved performance during inference across a wide range of DNNs, compared to standard \textsf{\small Adam}. Thus, integrating the NAG technique into optimizers proves to be highly beneficial.

Building on these insights, we propose a new optimizer, \textsf{\small SoftSignSGD (S3)}. The detailed implementation of \textsf{\small S3} is presented in Algorithm 1.
%
%\begin{equation}
%\begin{aligned}
%&m_{t} = \beta m_{t-1} + (1-\beta)g_t, \\
%&s_{t} = \beta v_{t-1} + (1-\beta)|g_t|^p, \\
%&d_{t} =\beta m_{t} + (1-\beta)g_t,  \\
%&b_{t} =\beta s_{t} + (1-\beta)|g_t|^p,  \\
%&x_{t+1} = x_t - \gamma_t \frac{m_{t}}{(b_t)^{\sfrac{1}{p}}},
%\end{aligned}
%\label{Eq.S3}
%\end{equation}

%where $p \ge 1$.

Key characteristics of \textsf{\small S3} are summarized below:

\emph{First}, \textbf{\textsf{\small S3} features a more general sign-like formulation with a flexible $p$-order momentum, departing from the conventional second-order momentum originally proposed for \textsf{\small Adam}.} As shown in Theorem \ref{theorem_2}, a larger $p$-order momentum allows the use of a larger learning rate, which promotes faster convergence and improved performance \cite{Large_learnign_rate_2020}. Additionally, training DNNs, especially large models, often results in abrupt changes in some gradient coordinates, which can destabilize training or even cause divergence without mitigation. In such instances, as demonstrated in Theorem \ref{theorem_2}, the update $\frac{|\bm{n}_t|}{\bm{b}_t(p)}$ in \textsf{\small S3} becomes smaller with larger $p$-order momentum, helping stabilize training by counteracting drastic gradient fluctuations. Figure 1 illustrates this behavior. Additionally, when training large models, computational cost becomes significant, and setting $p=1$ in \textsf{\small S3} only requires a computationally light element-wise absolute operation, minimizing computational overhead.

\emph{Second}, \textbf{\textsf{\small S3} uses the same exponential moving average coefficient $\beta$ for both $\bm{n}_t$ and $\bm{b}_t$, offering advantages in minimizing the risk of loss spikes and reducing the need for hyperparameter tuning.} As shown in Theorem \ref{theorem_2}, using the same $\beta$ ensures that the maximum value of each coordinate in the update $\frac{|\bm{n}_t|}{\bm{b}_t(p)}$ is minimized by 1. This design, as discussed in Section 2, helps mitigate the problem of loss spikes. In practice, this approach reduces tuning effort by removing one hyperparameter and lowers computational costs by eliminating the bias correction, which are required for \textsf{\small Adam}.

\emph{Third}, \textbf{\textsf{\small S3} introduces the NAG technique to accelerate training without increasing memory costs.} While previous works, such as \textsf{\small NAdam} and \textsf{\small Adan}, have integrated NAG into adaptive optimizers, \textsf{\small S3} distinguishes itself in several ways. In \textsf{\small S3}, NAG is applied to both the numerator and denominator of the update. The Nesterov momentum estimators in \textsf{\small S3} follow the NAG (II) formulation, which, as shown in Theorem \ref{theorem_3}, is equivalent to the vanilla NAG (I). The key advantage of this formulation is that \textsf{\small S3} does not require additional memory. In contrast, \textsf{\small NAdam} \cite{NAdam2016} only applies Nesterov momentum to the numerator and relies on complex bias-correction operations to stabilize training. Similarly, \textsf{\small Adan} \cite{Adan2024} employs Nesterov momentum in both the numerator and denominator, but its formulation, akin to NAG (III), requires more memory to store the previous gradient $\bm{g}_{t-1}$ and the new momentum $\bm{r}_k$ compared to \textsf{\small Adam}. Therefore, \textsf{\small Adan} may not be ideal for training large language models (LLMs) due to its higher memory requirements. Additionally, \textsf{\small Adan} introduces an additional momentum coefficient, increasing the tuning burden.

 \renewcommand\arraystretch{1.15}
 \begin{table*}[!tb]
  \centering
  \small
  \begin{threeparttable}
  \begin{tabular}{p{11cm}}
  \tabucline[1pt]{-}
  \textbf{Algorithm 1.}  \textsf{SoftSignSGD (S3)}    \\
  \tabucline[0.6pt]{-}
  \small
  \rownumber: \textbf{Input}:  the momentum  $\bm{m}_0=0$ ,  $\bm{s}_0=0$, the exponential moving average coefficient $\beta$ within $[0,1]$, the power factor $p$ within $[1, +\infty)$, and the learning rate sequence $\{\gamma_t\}$. \\
  \rownumber: \textbf{for} $t=1,...,T$ \textbf{do} \\
  \rownumber: \hspace{8pt} Randomly sample data and compute the gradient: $\bm{g}_t \leftarrow \nabla F(\bm{x}_{t};\zeta_t)$\\
  \rownumber: \hspace{8pt} Update the momentum $\bm{m}_t$: $\bm{m}_t \leftarrow \beta \bm{m}_{t-1}+ (1-\beta) \bm{g}_t $ \\
  \rownumber: \hspace{8pt} Update the momentum  $\bm{s}_t(p)$: $\bm{s}_t(p) \leftarrow \beta \bm{s}_{t-1} + (1-\beta) |\bm{g}_t|^p$ \\
  \rownumber: \hspace{8pt} Compute the Nesterov momentum $\bm{n}_t$: $\bm{n}_t \leftarrow \beta \bm{m}_{t}+ (1-\beta) \bm{g}_t $ \\
  \rownumber: \hspace{8pt} Compute the Nesterov momentum $\bm{b}_t(p)$: $\bm{b}_t(p) \leftarrow (\beta \bm{s}_t(p) + (1-\beta) |\bm{g}_t|^p)^{\sfrac{1}{p}}$ \\
  \rownumber: \hspace{8pt} Update the model parameter: $\bm{x}_{t+1} \leftarrow \bm{x}_t - \gamma_t\frac{\bm{n}_t}{\bm{b}_t(p)}$  \\
  \rownumber: \textbf{end for} \\
  \tabucline[1pt]{--}
  \end{tabular}
  \end{threeparttable}
  \label{Tab.1}
  \end{table*}

\begin{theorem}
The sequences $\{\bm{n}_t\}$ and $\bm{b}_t(p)$ are generated \textsf{\small S3} in Algorithm 1.  If the moving average coefficients for $\bm{m}_t, \bm{n}_t$ and $\bm{s}_t, \bm{b}_t$ of ar$\beta_1$ and $\beta_2$ which satisfy $\beta_1 <
\beta_2^{\sfrac{1}{p}}$ and $p \ge 1$,   it holds that

(1). The upper bound of  each element of the update $\frac{\bm{n}_t^{(j)}}{\bm{b}_t^{(j)}}$ is

\begin{equation}
\vspace{-5pt}
\frac{\vert\bm{n}_t^{(j)}\vert}{\bm{b}_t^{(j)}(p)} \le \frac{(1-\beta_1)}{(1-\beta_2)^{\sfrac{1}{p}}\left(1 - \frac{\beta_1^q}{\beta_2^{\sfrac{q}{p}}}\right)^{\sfrac{1}{q}}},
\vspace{-5pt}
\end{equation}
where $\frac{1}{p}+\frac{1}{q}=1$.

(2). When $\beta_1=\beta_2$, the  upper bound of  each element of the update $\frac{\bm{n}_t^{(j)}}{\bm{b}_t^{(j)}}$ reaches to the smallest $1$, i.e., $\frac{\vert\bm{n}_t^{(j)}\vert}{\bm{b}_t^{(j)}(p)} \le 1.$

(3). Let $1 \le p_1 \le p_2$, and then $\bm{b}_t(p_1) \le \bm{b}_t(p_2)$.
\label{theorem_2}
\end{theorem}

\begin{theorem}
The three formulations of NAG are listed in the following. Let $ \bm{x}_t=\tilde{ \bm{x}}_t-\gamma\beta \bm{m}_{t-1}$, the three formulations are equivalent, i.e.,
{\small
\begin{align}
\vspace{-5pt}
&\rm{NAG\;(I)}:  \left\{
\begin{aligned}
&  \bm{g}_t = \nabla f(\tilde{ \bm{x}}_t-\gamma\beta  \bm{m}_{t-1};\zeta_t ) \\
&  \bm{m}_t = \beta  \bm{m}_{t-1} + (1-\beta) \bm{g}_t \\
& \tilde{ \bm{x}}_{t+1} = \tilde{ \bm{x}}_t - \gamma  \bm{m}_t
\end{aligned}
 \right., \\
&\rm{NAG\;(II)}:  \left\{
\begin{aligned}
&  \bm{g}_t = \nabla f( \bm{x}_t;\zeta_t ) \\
&  \bm{m}_t = \beta  \bm{m}_{t-1} + (1-\beta)\bm{g}_t \\
& { \bm{x}}_{t+1} = { \bm{x}}_t - \gamma (\beta  \bm{m}_t + (1-\beta)  \bm{g}_t)
\end{aligned}
 \right., \\
&\rm{NAG\;(III)}:  \left\{
\begin{aligned}
&  \bm{g}_t = \nabla f( \bm{x}_t;\zeta_t ) \\
&  \bm{m}_t = \beta  \bm{m}_{t-1} + (1-\beta) \bm{g}_t \\
&  \bm{r}_t = \beta  \bm{r}_{t-1} + (1-\beta)( \bm{g}_t -  \bm{g}_{t-1}) \\
& {\bm{x}}_{t+1} = { \bm{x}}_t - \gamma (  \bm{m}_t + \beta \bm{r}_t)
\end{aligned}
 \right..
\end{align}
}

Moreover, if $\tilde{ {x}}_{t+1} \rightarrow \tilde{ {x}}_{t}$  as $ {m}_t\rightarrow 0$, ${ {x}}_{t} $ will  converge to $\tilde{ {x}}_{t} $.
\label{theorem_3}
\end{theorem}

\section{Theoretical Convergence Analysis}

To present the theoretical convergence guarantee  for \textsf{\small S3} (Algorithm 1) to optimize the nonconvex problem in Eq. (\ref{Eq.obj}),  we first introduce some necessary assumptions.

\textbf{Assumption 1.} [Bounded infimum] \emph{ There exists a constant $F^*$, the objective function follows $F(\bm{x}) \ge F^* $ for any $\bm{x} \in {\mathbb R}^d $. }

\textbf{Assumption 2. } [Generalized Smoothness] \emph{There exist constants $L_0, L_1, R\ge 0$,
for any $\bm{x}, \bm{y} \in {\mathbb R}^d $ with $\Vert \bm{x} - \bm{y}\Vert_2 \le R$ , the objective function follows,}
\begin{equation}
 \Vert \nabla F(\bm{y}) - \nabla F(\bm{x}) \Vert_2 \le (L_0 + L_1 \Vert \nabla F(\bm{x}) \Vert_2) \Vert \bm{x} - \bm{y} \Vert_2.
\end{equation}

\textbf{Assumption 3. }[Unbias noisy  gradient and bounded variance] \emph{There exists a constant $\sigma$. For  $\bm{x}_t\in {\mathbb R}^d $ at any time, the noisy gradient of the objective function obeys follows}

\begin{equation}
 {\mathbb E} [\bm{g}_t] ={\mathbb E} [\nabla f(\bm{x}_t;\zeta_t)]= \nabla F(\bm{x}_t), ~~~~~~ {\mathbb E} [\Vert \bm{g}_t - \nabla F(\bm{x}_t) \Vert_2^2] \le \sigma^2.
\end{equation}

%%\textbf{Assumption 4. }[Bounded gradient] \emph{The noisy gradient and the full-set gradient are bounded i.e.,
%% $\Vert g_t\Vert_2 \le G, ~~~ \Vert \nabla F(x_t) \Vert_2 \le G, ~~~~ \forall t\ge1.$
% }

Under the assumptions above, we then present the theoretical convergence for \textsf{\small S3} in Theorem  \ref{convergence}.

\begin{theorem}
$\{x_t\}_{t=1}^T$ is generated by  Algorithm 1 under Assumption 1-4. Let the hyperparameters be set as $\beta=1-\frac{1}{\sqrt{T}}$ and $\gamma=\frac{1}{L_0T^{3/4}}$.  If $u_t= \frac{\vert \bm{n}_t^{(j)} \vert}{\bm{b}_t^{(j)}} \ge \frac{1}{U_{\max}}$, then
\begin{equation}
\begin{aligned}
\frac{1}{T}\sum_{t=1}^T \mathbb E [\Vert\nabla F(\bm{x}_t)\Vert_1] \le& \frac{2L_0U_{\max}(F(\bm{x}_{1})- F(\bm{x}^*))}{T^{\sfrac{1}{4}}} \\
&+ \frac{4\beta U_{\max}\sqrt{d}{\mathbb E}\left[\left\Vert \nabla F(\bm{x}_1)\right\Vert_2\right]}{T^{\sfrac{1}{2}}} \\
&+\frac{{4}U_{\max}\sqrt{d}\sigma}{T^{\sfrac{1}{4}}} +  \frac{4\beta^2U_{\max} {d}}{T^{\sfrac{1}{4}}} + \frac{U_{\max}d}{T^{\sfrac{7}{4}}}.
\end{aligned}
\end{equation}
\label{convergence}
\vspace{-10pt}
\end{theorem}

\textbf{Remark 1 [Adopting Weaker Assumption].} The theoretical convergence analysis for \textsf{\small S3} in Theorem \ref{convergence} relies on a general non-uniform smoothness condition (Assumption 2), which is weaker than the assumptions required in previous works that analyzed the convergence of \textsf{\small Adam}. For instance, \cite{Adam_type_proof_2018, Adam_simple_2020} proved convergence for non-convex objectives under the assumption that gradients are bounded. \cite{Nesterov_Adam_2018} required the gradients to maintain consistent signs along the trajectory, even when considering Nesterov acceleration. \cite{Adam_proof_2022} assumed uniform $L$-smoothness but only proved convergence to a neighborhood of stationary points with a fixed radius. More recently, \cite{Adam_proof_weak_assumption_2023, Adam_proof_weak_assumption_2024} provided convergence bounds for \textsf{\small Adam} under the general non-uniform smoothness assumption, but the convergence was only in probability.

\textbf{Remark 2 [Achieving Optimal Convergence Rate].} As demonstrated in Theorem \ref{convergence}, the convergence rate of \textsf{\small S3} is $O\left(\frac{1}{T^{\sfrac{1}{4}}}\right)$, which matches the established lower bound for optimal convergence in non-convex stochastic optimization \cite{SGD_lower_bound_2023}.

 \begin{figure*}[!th]
 \centering
  \subfloat[\footnotesize{ResNet-50, Train Loss}]{\includegraphics[width=.42\textwidth]{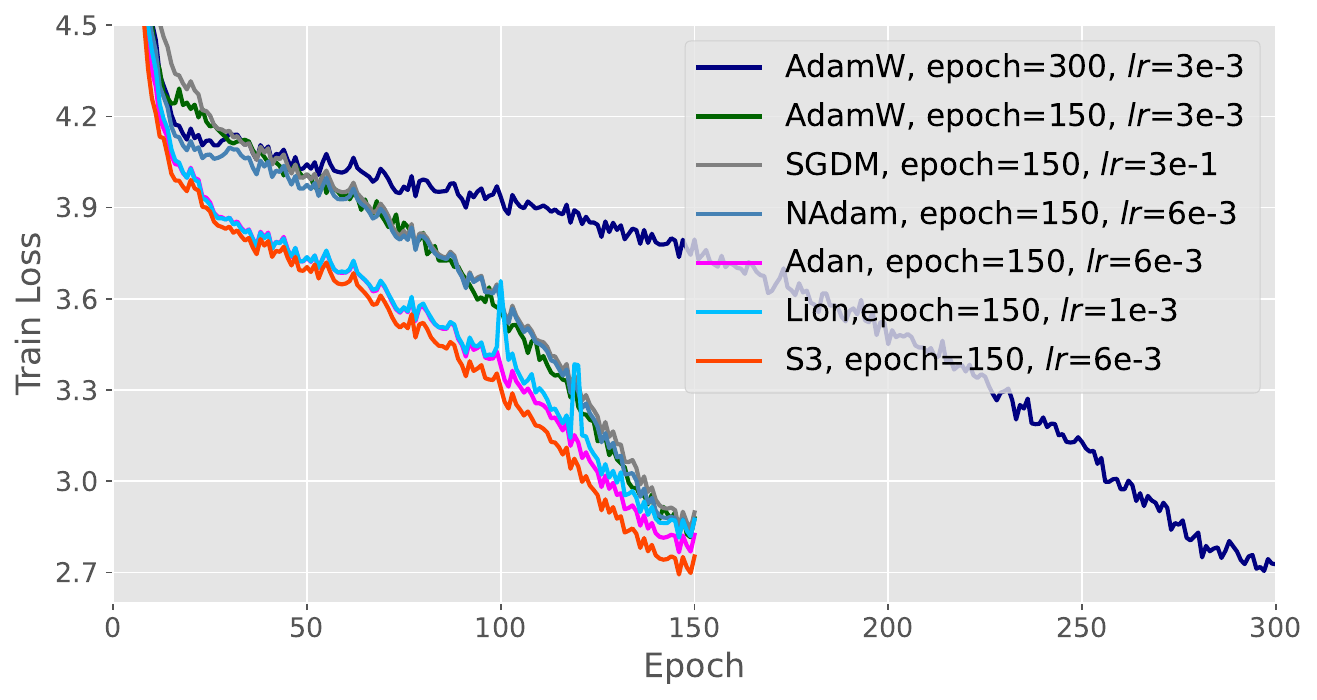}} \hspace{20 pt}
  \subfloat[\footnotesize{ResNet-50, Test Accuracy} ]{\includegraphics[width=.42\textwidth]{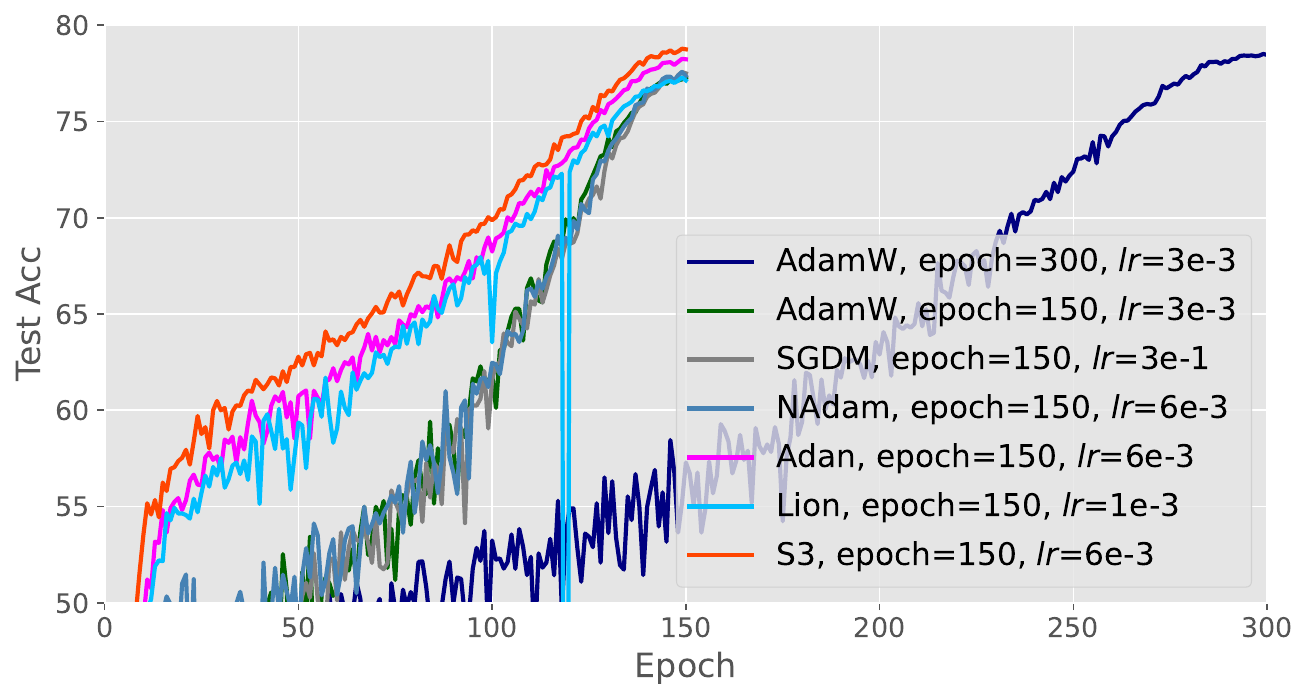}} \\
  \subfloat[\footnotesize{ViT-B/16, Train Loss}]{\includegraphics[width=.42\textwidth]{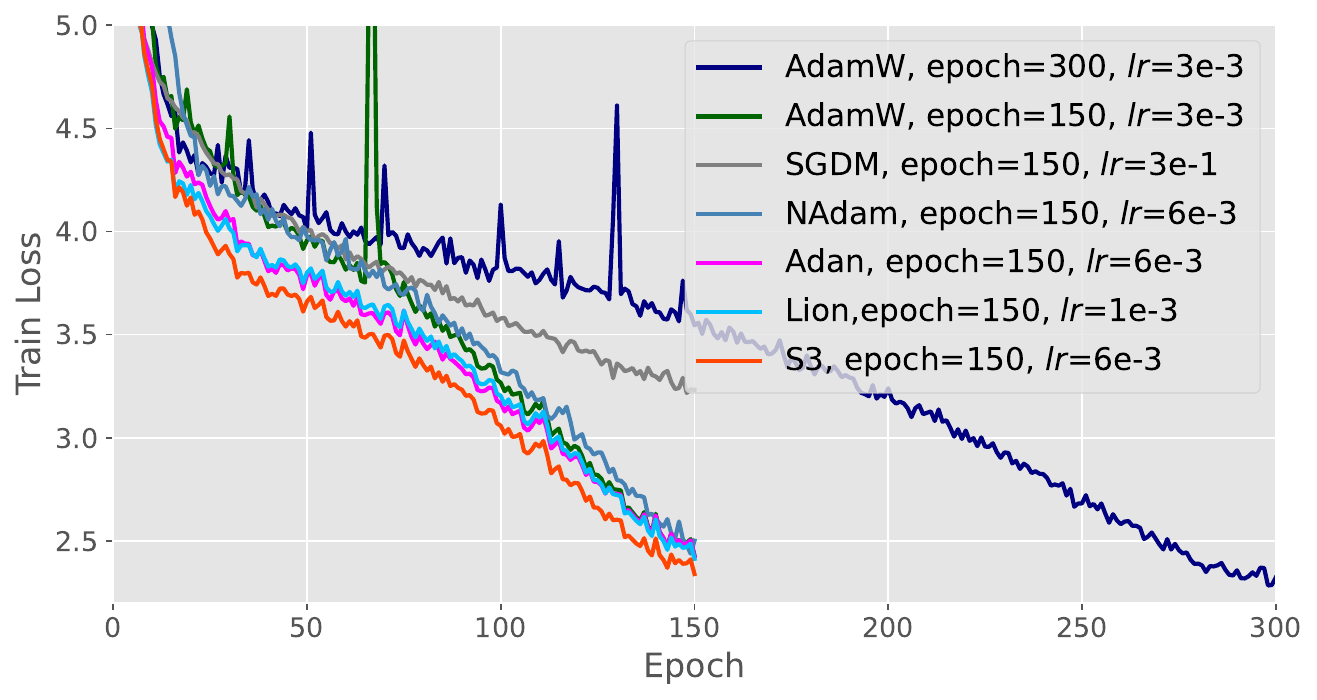}} \hspace{20 pt}
  \subfloat[\footnotesize{ViT-B/16, Test Accuracy} ]{\includegraphics[width=.42\textwidth]{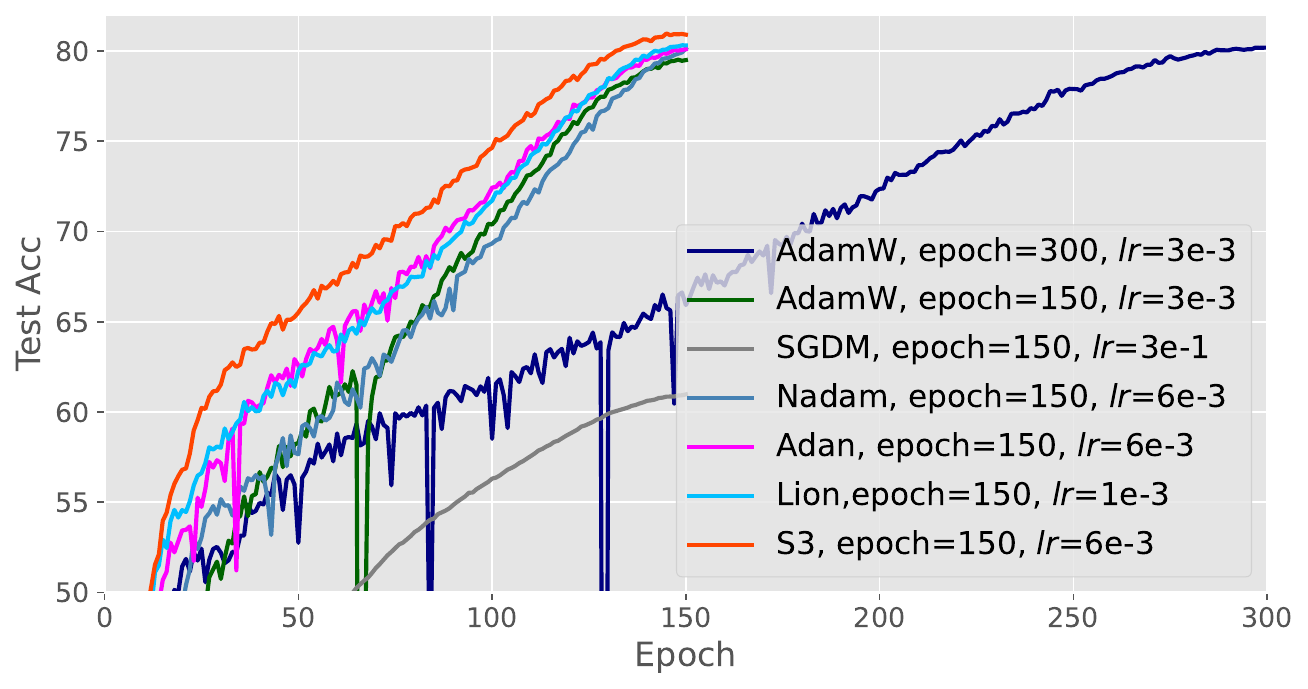}}
   \vspace{-5pt}
  \caption{\small{ Comparison of train loss and test accuracy  on ImageNet for training ReNet-50 and ViT-B/16 with \textsf{\footnotesize AdamW}, \textsf{\footnotesize SGDM}, \textsf{\footnotesize NAadm}, \textsf{\footnotesize Adan}, \textsf{\footnotesize Lion} and \textsf{\footnotesize S3}. } }
  \label{Fig.3}
\end{figure*}

\begin{figure*}[!tb]
 \centering
  \subfloat[\footnotesize{GPT-2 (345M), Train Loss}]{\includegraphics[width=.42\textwidth]{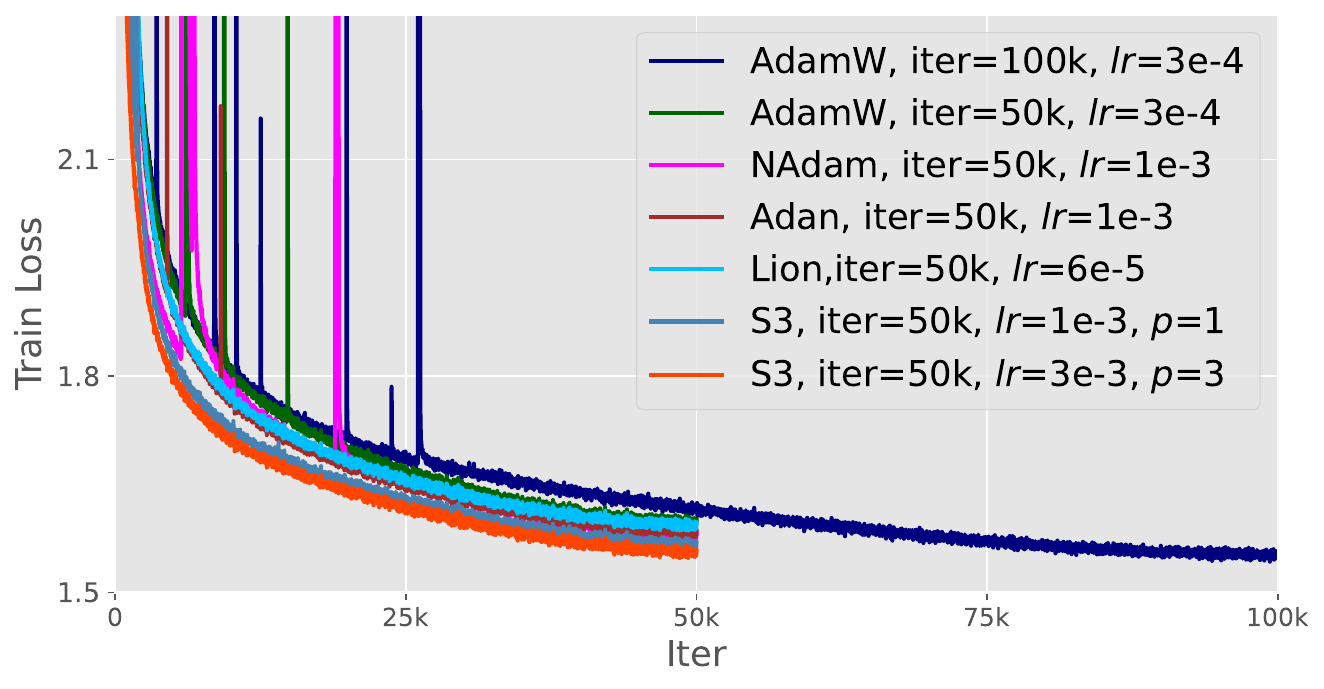}} \hspace{20 pt}
  \subfloat[\footnotesize{GPT-2 (345M), Validation Loss} ]{\includegraphics[width=.42\textwidth]{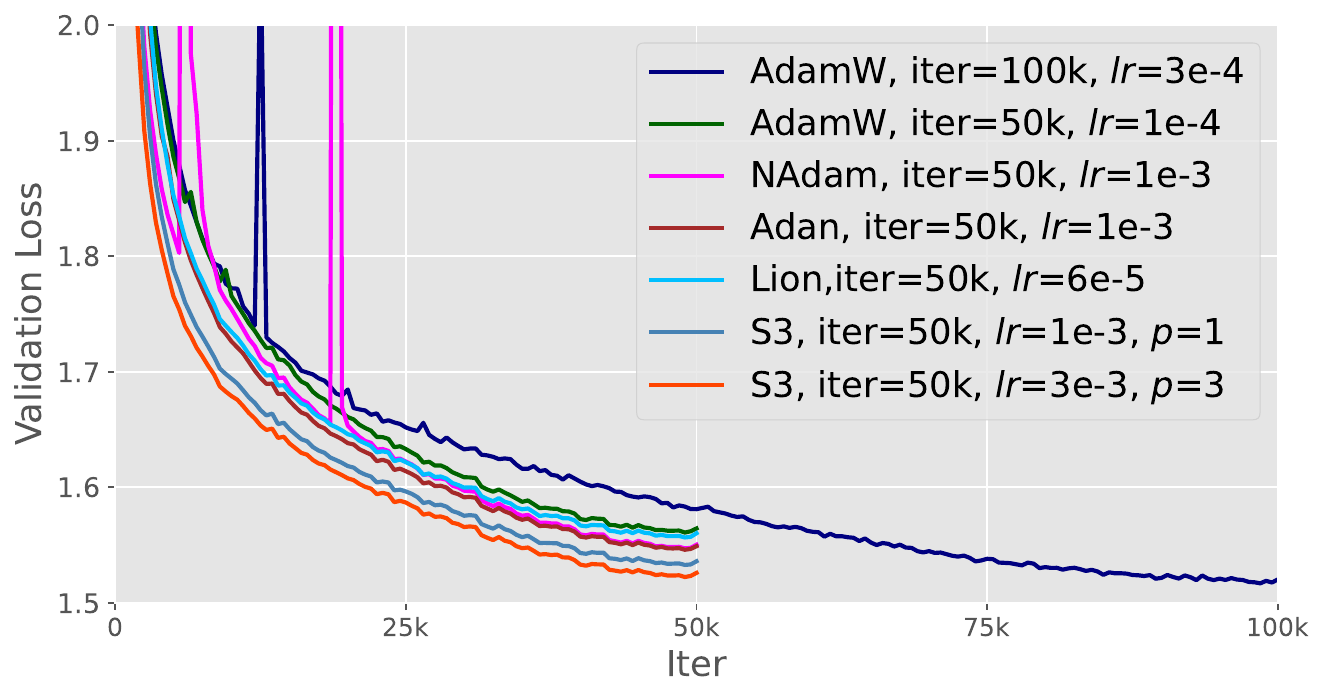}} \\
  \subfloat[\footnotesize{GPT-2 (7B), Train Loss}]{\includegraphics[width=.42\textwidth]{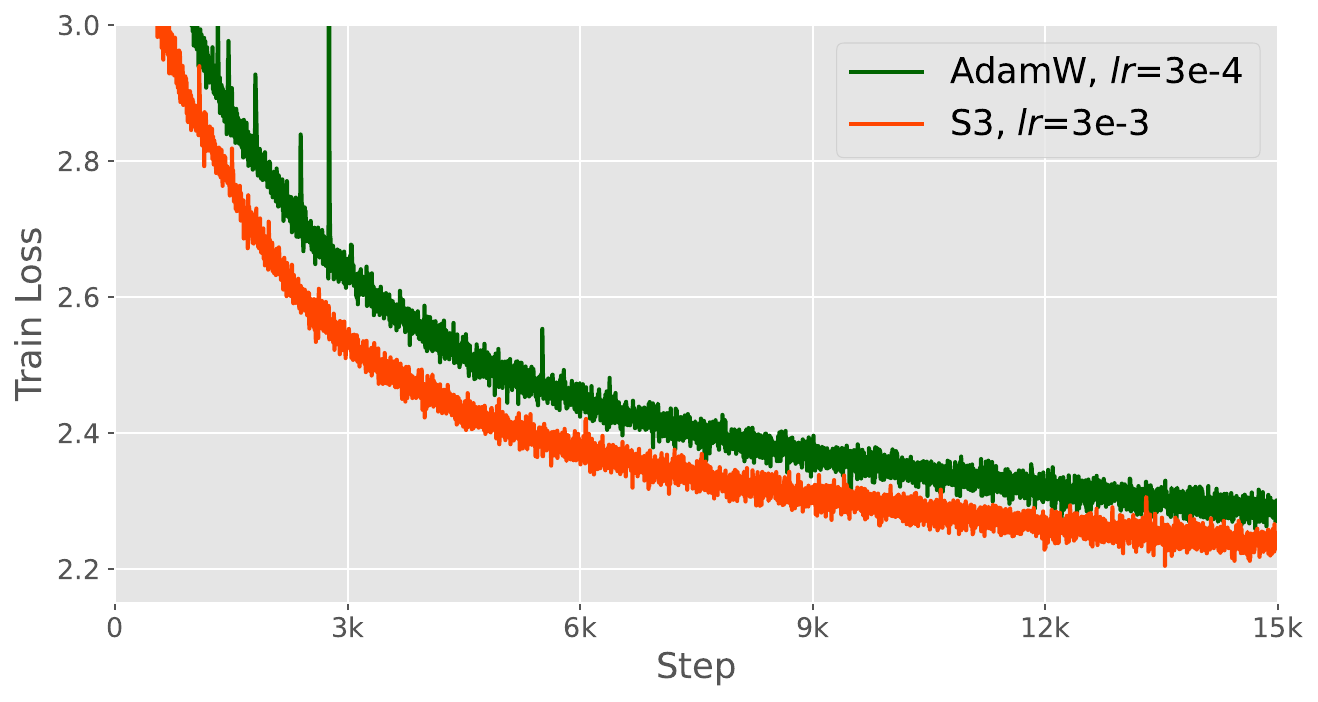}} \hspace{20 pt}
  \subfloat[\footnotesize{GPT-2 (7B), Validation Loss} ]{\includegraphics[width=.42\textwidth]{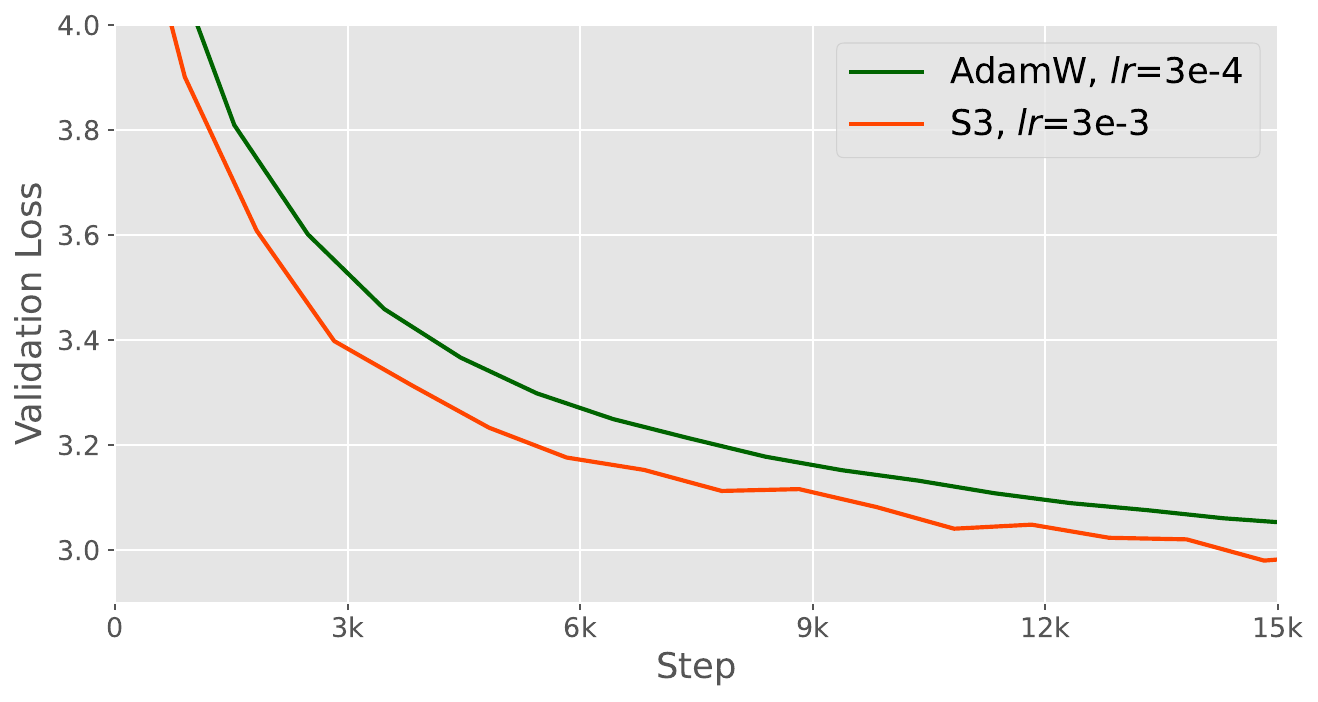}}
  \caption{{ Comparison of train loss and validation loss  for pre-training GPT-2 (345M) and GPT-2 (7B) with \textsf{\footnotesize AdamW}, \textsf{\footnotesize NAdam},\textsf{\footnotesize Adan}, \textsf{\footnotesize Lion} and \textsf{\footnotesize S3}. } }
  \label{Fig.5}
\end{figure*}

\section{Experiment}

We compare \textsf{\small S3} with several representative optimizers, including \textsf{\small SGDM} \cite{SGD_1951}, \textsf{\small AdamW} \cite{AdamW2017}, \textsf{\small NAdam} \cite{NAdam2016}, \textsf{\small Adan} \cite{Adan2024}, and \textsf{\small Lion} \cite{Lion2023}, across both vision and language tasks. For vision tasks, we evaluate classification accuracy by training the CNN-based ResNet-50 \cite{ResNet2016} and the Transformer-based ViT-Base \cite{ViT_2020} on ImageNet. In language tasks, we assess pre-training performance by training GPT-2 (345M) and GPT-2 (7B) \cite{GPT2_2019} on OpenWebText and the refined CommonCrawl datasets.

 \renewcommand\arraystretch{1.1}
\begin{table*}[!hptb]
\small
\centering
\caption{ {  Test accuracy  (\%)  on ImageNet for training ResNet-50 and ViT-B/16 with \textsf{\footnotesize AdamW}, \textsf{\footnotesize SGDM}, \textsf{\footnotesize Adan}, \textsf{\footnotesize Lion} and \textsf{\footnotesize S3}.} }
\vspace{0pt}
\begin{tabu}{ |l| p{1.3cm}<{\centering} p{1.3cm}<{\centering} p{1.3cm}<{\centering}  p{1.3cm}<{\centering} p{1.3cm}<{\centering}  p{1.3cm}<{\centering} | p{1.5cm}<{\centering}| }
%\hline
\tabucline[1pt]{-}
\multirow{2}{*}{Network}    &\multicolumn{6}{c|}{150 epochs} &300 epochs \\
 & \textsf{\footnotesize AdamW}     &\textsf{\footnotesize SGDM} &\textsf{\footnotesize NAdam}   &\textsf{\footnotesize Adan}  &\textsf{\footnotesize Lion} &\textsf{\footnotesize S3}  &\textsf\footnotesize {AdamW}\\

\hline
\hline
ResNet-50           & 77.29  &77.50  &77.36  &{78.23}   &{77.14}   &\textbf{78.76}  &78.46     \\
%\hline
ViT-B/16            &{79.52} &{60.99} &80.31 &{80.11} &{80.32}  &\textbf{80.93}   &{80.13}   \\
%\hline
\tabucline[1pt]{-}
\end{tabu}
\label{Tab.1}
\end{table*}

\renewcommand\arraystretch{1.0}
\begin{table*}[!hptb]
\small
\centering
\caption{ { Validation perplexity (the lower, the better) for training GPT-2 (345M) and  GPT-2 (7B).} }
\vspace{0pt}
\begin{tabu}{ |l|l| p{1.05cm}<{\centering} p{1.05cm}<{\centering}  p{1.05cm}<{\centering}p{1.05cm}<{\centering}   p{1.5cm}<{\centering} | p{1.55cm}<{\centering}| }
%\hline
\tabucline[1pt]{-}
\multirow{3}{*}{Network} &\multirow{3}{*}{Dataset}   &\multicolumn{5}{c|}{50k steps}  &100k steps \\
& & \textsf{\footnotesize AdamW}  &\textsf{\footnotesize NAdam}      &\textsf{\footnotesize Adan}  &\textsf{\footnotesize Lion}  &\textsf{\footnotesize S3}  &\textsf\footnotesize {AdamW}\\  & & ({\scriptsize $lr$=3e-4})   & ({\scriptsize$lr$=3e-4})   &({\scriptsize $lr$=1e-3})      &({\scriptsize $lr$=6e-5})  &({\scriptsize$p$=3,$lr$=3e-3})   &({\scriptsize $lr$=3e-4}) \\

\hline
\hline
GPT-2 (345M)   &{\footnotesize OpenWebText}       & 4.78 &4.71 &{4.69}   &{4.76}   &\textbf{4.59}  &4.57     \\
%\hline
GPT-2 (7B)     &{\footnotesize CommonCrawl}    &{21.13} &- &- &- &\textbf{19.69}   &-   \\
%\hline
\tabucline[1pt]{-}
\end{tabu}
\label{Tab.2}
\end{table*}

 \begin{figure*}[ht!]
   \vspace{-5pt}
 \centering
  \subfloat[\footnotesize{GPT-2 (345M)}]{\includegraphics[width=.38\textwidth]{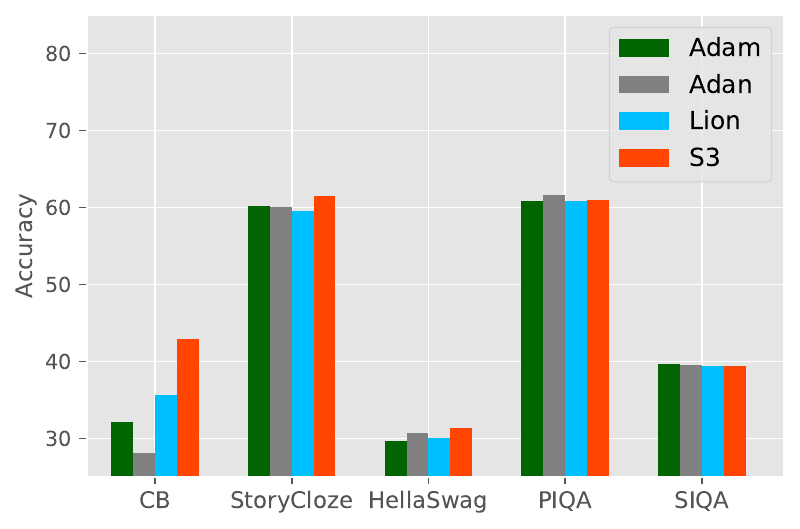}} \hspace{50 pt}
  \subfloat[\footnotesize{GPT-2 (7B)} ]{\includegraphics[width=.38\textwidth]{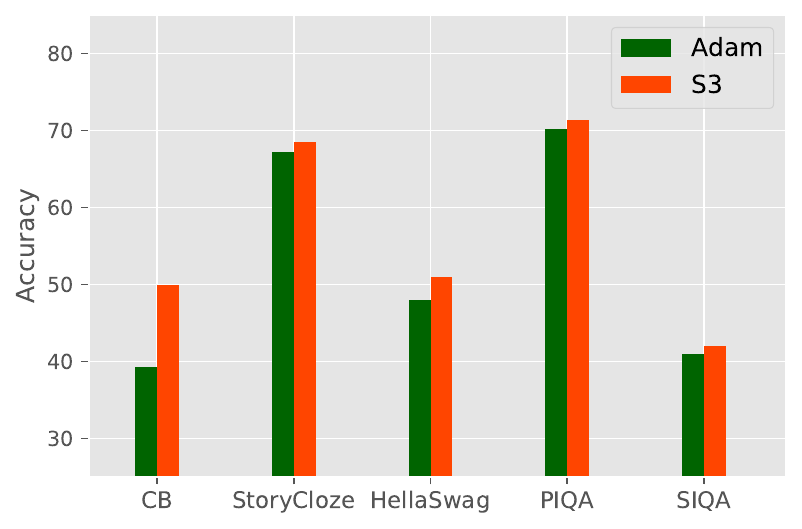}} \\
   \vspace{-5pt}
  \caption{\small{ Zero-shot  evaluation of the pre-trained GPT-2 (345M) and  GPT-2 (7B) with \textsf{\footnotesize AdamW}, \textsf{\footnotesize Adan}, \textsf{\footnotesize aLion}, and \textsf{\footnotesize S3} on downstream reasoning tasks. } }
  \label{Fig.10}
  \vspace{-5pt}
\end{figure*}

 \subsection{Training Setting}

We use the PyTorch vision reference code\footnote{https://github.com/pytorch/vision/tree/main/references/classification} to implement the vision tasks. For data augmentation, we follow the recommended settings in the reference code, incorporating RepeatedAugment, AutoAugment Policy (with magnitude set to $9$), and Mixup ($0.2$)/CutMix ($1.0$) with a probability of $0.5$. Additionally, we apply label smoothing with a value of $0.11$. The batch size is set to $1024$ for ResNet-50 and $4096$ for ViT-B/16. For the learning rate scheme, we linearly increase it to its peak over the initial $30$ epochs, followed by a cosine decay, decreasing it to $0$ in the subsequent epochs.
The customized hyperparameters for \textsf{\small SGD} and \textsf{\small AdamW} are set according to well-established practices in the reference code. For \textsf{\small Adan} and \textsf{\small Lion}, we follow the recommendations provided in their respective official papers \cite{Adan2024, Lion2023} for training ResNet-50 and ViT-B/16. Since \textsf{\small NAdam} is similar to \textsf{\small AdamW}, we use the same hyperparameters for both optimizers.

For the language tasks, we implement them using Megatron-LM\footnote{https://github.com/NVIDIA/Megatron-LM}. Following the standard GPT-2 protocol, we construct a 345M Transformer decode-only model with $12$ layers, a sequence length of $1024$, a hidden size of $512$, and $8$ attention heads. To test the effectiveness of \textsf{\small S3} in training large language models, we also construct a 7B model with $32$ layers, a sequence length of $4096$, a hidden size of $4096$, and $32$ attention heads. GPT-2 (345M) is trained on OpenWebText with a batch size of $512$, and GPT-2 (7B) is trained on the refined CommonCrawl dataset with a batch size of $1024$. For \textsf{\small S3}, we set $p=3$ and do not employ gradient clipping. For other optimizers, the gradient clipping threshold is set to $1.0$.
For the learning rate scheme, we linearly increase it to its peak over the first $5k$ steps, then decrease it to $0.1\times$ of the peak using cosine decay in the subsequent steps. The peak learning rates for all optimizers are determined through coarse search for training GPT-2 (345M).

 \subsection{Experiments for Vision Tasks}

Compared to the baseline \textsf{\small AdamW}, Figures \ref{Fig.3}(a) and \ref{Fig.3}(c) show that \textsf{\small S3} exhibits significantly faster convergence and achieves a notably smaller final training loss. As depicted in Figures \ref{Fig.3}(b) and \ref{Fig.3}(d), as well as in Table \ref{Tab.1}, \textsf{\small S3} achieves test accuracies that are $1.47\%$ and $1.36\%$ higher when training ResNet-50 and ViT-B, respectively. This represents a substantial improvement in both training speed and inference accuracy. Even when the training epochs for \textsf{\small AdamW} are increased by $2\times$, \textsf{\small S3} remains comparable and even slightly outperforms \textsf{\small AdamW}. Furthermore, while \textsf{\small AdamW} experiences loss spikes during the training of ViT-B, \textsf{\small S3} maintains training stability, even with a large learning rate.

Additionally, other competitive optimizers, including \textsf{\small SGDM}, \textsf{\small Adan}, and \textsf{\small Lion}, were also evaluated and their results are presented in Figure \ref{Fig.3} and Table \ref{Tab.1}. While \textsf{\small SGDM} performs similarly to \textsf{\small AdamW} on the CNN-based ResNet-50, it underperforms on the Transformer-based ViT-B, consistent with the analysis in Section 2. The introduction of the NAG technique in \textsf{\small Adan} and \textsf{\small Lion} leads to improvements in training speed and test accuracy over \textsf{\small AdamW}, but they still lag behind \textsf{\small S3}. It is also worth noting that both \textsf{\small Adan} and \textsf{\small Lion} encounter stability issues during training.

\subsection{Experiments for Language Tasks}

As shown in Figure \ref{Fig.5} and Table \ref{Tab.2}, \textsf{\small S3} consistently achieves faster training convergence and lower validation perplexity compared to \textsf{\small AdamW}. The superiority of \textsf{\small S3} becomes even more apparent as the model size increases. Notably, the improvement observed in the 345M model with \textsf{\small S3} is comparable to the performance achieved by \textsf{\small AdamW} with twice the number of steps. This translates into a significant reduction in the number of steps and total computation required to reach the same loss level, resulting in substantial time and cost savings for LLM pre-training. Furthermore, while \textsf{\small AdamW} frequently experiences loss spikes, \textsf{\small S3} rarely encounters this issue, even with a learning rate that is $10\times$ larger. Additionally, the large $p$-order momentum in \textsf{\small S3} facilitates the use of a larger learning rate, further accelerating training and improving performance.

In addition, \textsf{\small Lion} and \textsf{\small Adan} were also evaluated on the GPT-2 (345M) model. Although \textsf{\small Lion} converges faster than \textsf{\small AdamW} initially, it does not exhibit superiority in final validation perplexity. \textsf{\small Adan} slightly outperforms \textsf{\small AdamW} in both training speed and validation performance. However, as discussed in Section 3, \textsf{\small Adan} requires more memory and additional hyperparameter tuning, making it less appealing for LLM pre-training. Additionally, \textsf{\small Adan} is also prone to experiencing loss spikes.

To further validate the effectiveness of the proposed optimizers, we conducted evaluation experiments on pre-trained GPT-2 models, specifically GPT-2 (345M) and GPT-2 (7B), using downstream reasoning benchmarks from OpenCompass\footnote{https://github.com/open-compass/opencompass}. As depicted in Figure \ref{Fig.10}, the results demonstrate that \textsf{\small S3} consistently outperforms \textsf{\small Adam} across the majority of benchmarks. This superiority is evident in the improved downstream accuracy, indicating that the lower validation loss achieved by \textsf{\small S3} translates into enhanced performance on these reasoning tasks.

An interesting observation is that the superiority of \textsf{\small S3} becomes more pronounced as the model size becomes large. This suggests that the benefits of \textsf{\small S3} extend beyond its effectiveness with smaller models, showcasing its scalability and adaptability to larger and more complex architectures.

It is essential to acknowledge that the GPT-2 (345M) model, along with its training dataset, is relatively small. Consequently, the pre-trained models may lack the inherent capabilities needed to perform well on downstream benchmarks, regardless of the optimizer used. Consequently, the accuracies achieved by  GPT-2 (345M) with these optimizers may exhibit a degree of randomness due to the model's inherent limitations in handling more complex tasks with a smaller scale.

\subsection{Ablation Study}

 \begin{figure}[h]
 \centering
 \begin{minipage}[t]{0.5\textwidth}
   \centering
  \subfloat{\includegraphics[width=.9\textwidth]{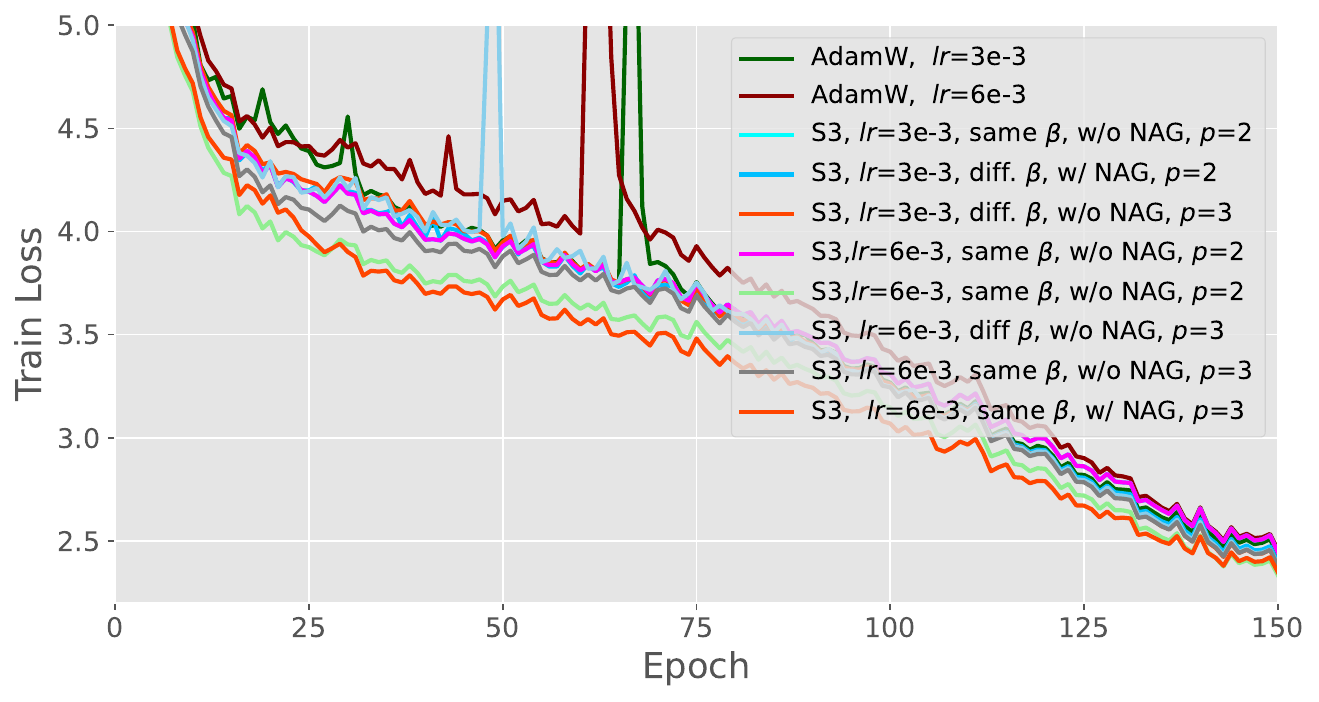}}
   \vspace{-5pt}
  \caption{\small{ Ablation study on train loss of \textsf{\footnotesize S3} for training ViT-B/16  on ImageNet. } }
    \vspace{5pt}
  \label{Fig.6}
  \end{minipage}

  \begin{minipage}[t]{0.5\textwidth}
   \centering
    \subfloat{\includegraphics[width=.9\textwidth]{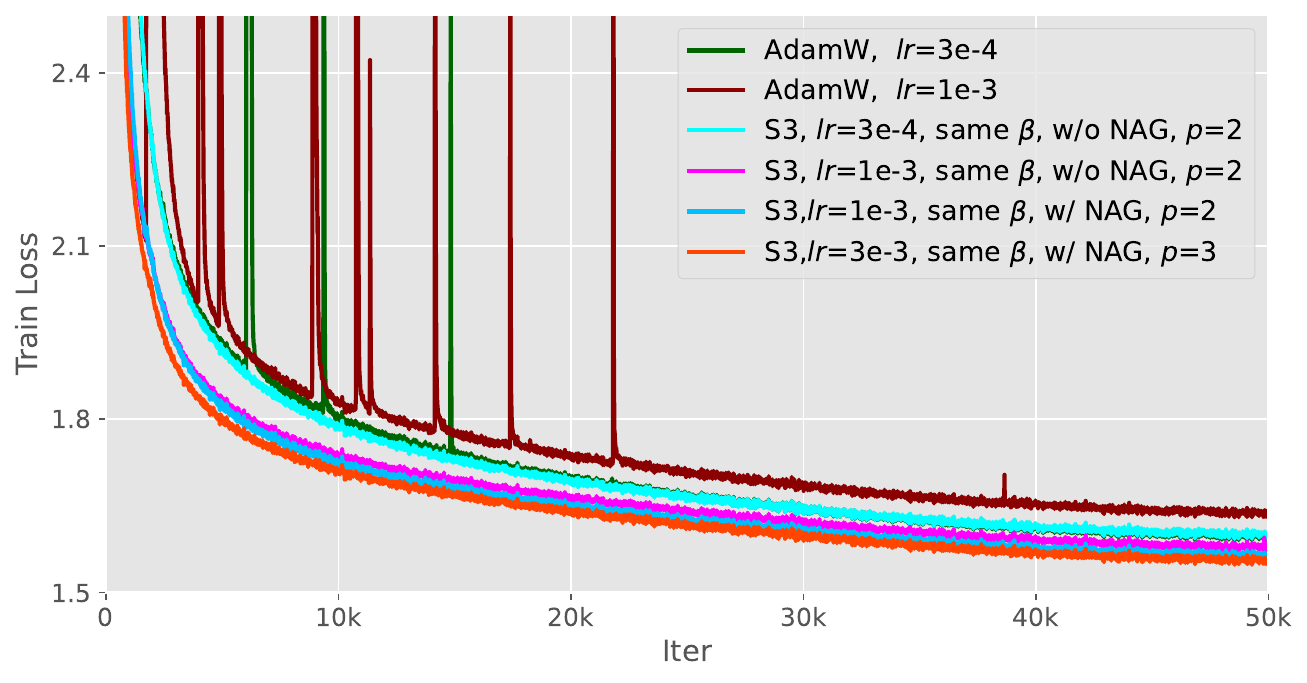}}
       \vspace{-5pt}
  \caption{\small{ Ablation study on train loss of \textsf{\footnotesize S3} for training GPT-2(345M)  on OpenWebText.} }
  \label{Fig.6_5}
  \end{minipage}

  \vspace{5pt}
 \begin{minipage}[t]{0.5\textwidth}
  \centering
  \subfloat{\includegraphics[width=.7\textwidth]{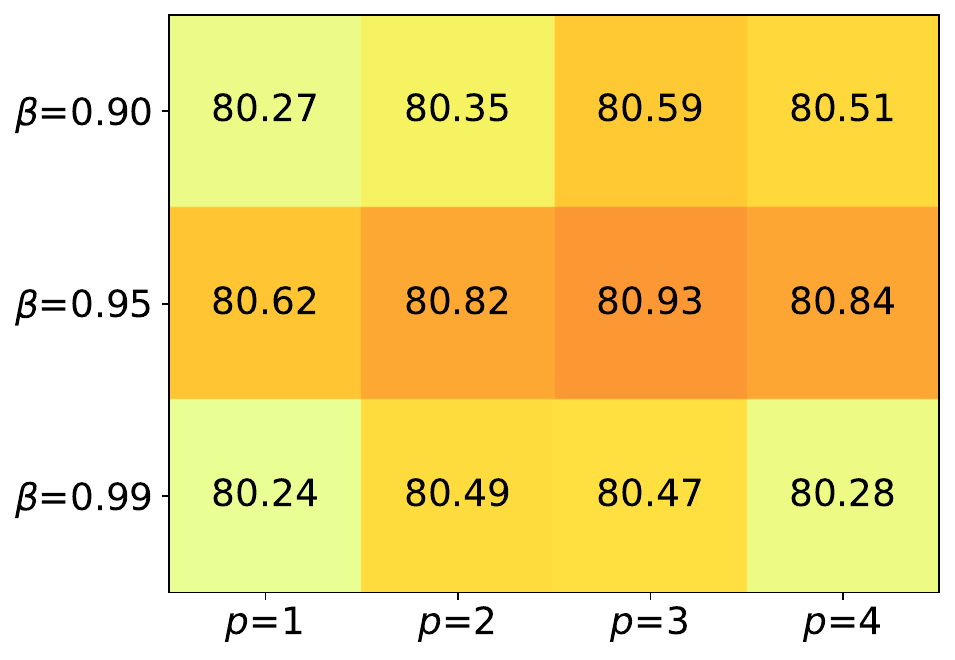}}
   \vspace{-5pt}
  \caption{\small{ Impact of the momentum order ($p$) and the momentum coefficient ($\beta$) on the Accuracy of \textsf{\footnotesize S3} training ViT-B/16 on ImageNet.  } }
  \label{Fig.7}
 \end{minipage}

\end{figure}

%\begin{figure}[h]
% \centering
%  \subfigure{\includegraphics[width=.30\textwidth]{./figures/sen_analysis_hotmap.pdf}}
%    \vspace{-10pt}
%  \caption{\small{ Impact of the momentum order ($p$) and the momentum coefficient ($\beta$) on the Accuracy of \textsf{\footnotesize S3} training ViT-B/16 on ImageNet.  } }
%  \label{Fig.7}
%  \vspace{-8pt}
%\end{figure}
We implement ablation experiments for training ViT-B/16 to clarify the contributions of each modification of \textsf{\small S3} over \textsf{\small Adam}.  Figure \ref{Fig.6} and Table \ref{Tab.3} showcase that employing a large learning rate and sharing the same $\beta$ alone have little or even a negative impact on performance (e.g., Exp. \ding{172} vs. Exp. \ding{173}, Exp. \ding{172} vs. Exp. \ding{174}), while their combination results in a notable improvement (e.g., Exp. \ding{173}, Exp. \ding{174} vs. Exp. \ding{177}, Exp. \ding{179} vs. Exp. \ding{180}). In contrast, harnessing a larger $p$ can have an individually beneficial effect on performance (e.g., Exp. \ding{172} vs. Exp. \ding{176}, Exp. \ding{173} vs. Exp. \ding{179}), and the performance gain from the benefits of NAG is more pronounced than other modification (e.g., Exp. \ding{172} vs. Exp. \ding{175}, Exp. \ding{177} vs. Exp. \ding{178}, Exp. \ding{180} vs. Exp. \ding{181}).

%Figure \ref{Fig.6} clearly showcases that sharing the exponential moving average coefficient $\beta$ for both the numerator momentum and the denominator momentum can mitigate the occurrence of loss spikes. However, "However, this enhancement does not directly impact training convergence speed and classification accuracy.  As shown in Tables \ref{Fig.6} and \ref{Tab.3}, employing a large $p$-order momentum for the denominator allows a larger learning rate, resulting in a smaller training loss and better inferencing performance. Furthermore, harnessing the NAG technique further accelerates the convergence speed and helps \textsf{\small S3} achieve a high classification accuracy.

\renewcommand\arraystretch{1.15}
\begin{table*}[!hptb]
\small
\centering
\caption{ Ablation study on test accuracies (\%) of \textsf{\footnotesize S3} for training  ViT-B/16.}
\begin{tabular}{|c| p{1.3cm}<{\centering} p{1.3cm}<{\centering} p{1.3cm}<{\centering} p{1.3cm}<{\centering} |l|}
%\begin{tabular}{|c c c|c|}
%\begin{tabular}{|l|c|}
\hline
 Exp. &large $lr$  &same $\beta$  & NAG &flexible $p$ &Test Accuracy   \\
\hline
\hline
{\normalsize \ding{192}} &- &- &- &-   &79.52  (\textsf{\footnotesize AdamW},$lr$=3e-3)  \\
{\normalsize \ding{193}} &\textcolor{cyan}{\CheckmarkBold} &-          &-          &-           &79.45  (\textsf{\footnotesize AdamW}, $lr$=6e-3)  \\
{\normalsize \ding{194}} &-          &\textcolor{cyan}{\CheckmarkBold} &-          &-           &79.48 (\textsf{\footnotesize S3}, $lr$=3e-3, same  $\beta$, w/o NAG, $p=2$) \\
{\normalsize \ding{195}} &-          &-          &\textcolor{cyan}{\CheckmarkBold} &-           &80.17 (\textsf{\footnotesize S3}, $lr$=3e-3, diff. $\beta$, w/ NAG,  $p=2$) \\
{\normalsize \ding{196}} &-          &-          &-          &\textcolor{cyan}{\CheckmarkBold}  &79.74 (\textsf{\footnotesize S3}, $lr$=3e-3, diff. $\beta$, w/o NAG, $p=3$) \\
{\normalsize \ding{197}} &\textcolor{cyan}{\CheckmarkBold} &\textcolor{cyan}{\CheckmarkBold} &-          &-           &80.25 (\textsf{\footnotesize S3}, $lr$=6e-3, same  $\beta$, w/o NAG, $p=2$) \\
{\normalsize \ding{198}} &\textcolor{cyan}{\CheckmarkBold} &\textcolor{cyan}{\CheckmarkBold} &\textcolor{cyan}{\CheckmarkBold} &-           &80.82 (\textsf{\footnotesize S3}, $lr$=6e-3, same  $\beta$, w/ NAG,  $p=2$) \\
{\normalsize \ding{199}} &\textcolor{cyan}{\CheckmarkBold} &-          &-          &\textcolor{cyan}{\CheckmarkBold}  &79.98 (\textsf{\footnotesize S3}, $lr$=6e-3, diff.  $\beta$, w/o NAG, $p=3$) \\
{\normalsize \ding{200}} &\textcolor{cyan}{\CheckmarkBold} &\textcolor{cyan}{\CheckmarkBold} &-          &\textcolor{cyan}{\CheckmarkBold}  &80.31 (\textsf{\footnotesize S3}, $lr$=6e-3, same  $\beta$, w/o NAG, $p=3$)\\
{\normalsize \ding{201}} &\textcolor{cyan}{\CheckmarkBold} &\textcolor{cyan}{\CheckmarkBold} &\textcolor{cyan}{\CheckmarkBold} &\textcolor{cyan}{\CheckmarkBold}  &80.93 (\textsf{\footnotesize S3}, $lr$=6e-3, same  $\beta$, w/ NAG,  $p=3$) \\
\hline
\end{tabular}
\label{Tab.3}
\end{table*}

\renewcommand\arraystretch{1.2}
\begin{table*}[!hptb]
\small
\centering
\caption{ Ablation study on validation perplexity  of \textsf{\footnotesize S3} for training  GPT-2(345M).}
\begin{tabular}{|c| p{1.3cm}<{\centering} p{1.3cm}<{\centering} p{1.3cm}<{\centering} p{1.3cm}<{\centering} |l|}
%\begin{tabular}{|c c c|c|}
%\begin{tabular}{|l|c|}
\hline
 Exp. &large $lr$  &same $\beta$  & NAG &flexible $p$ &Validation perplexity   \\
\hline
\hline
{\normalsize \ding{192}} &- &- &- &-   &4.78  (\textsf{\footnotesize AdamW},$lr$=3e-4)  \\
{\normalsize \ding{193}} &\textcolor{cyan}{\CheckmarkBold} &-          &-          &-           &4.97  (\textsf{\footnotesize AdamW}, $lr$=1e-3)  \\
{\normalsize \ding{194}} &-          &\textcolor{cyan}{\CheckmarkBold} &-          &-           &4.77 (\textsf{\footnotesize S3}, $lr$=3e-4, same  $\beta$, w/o NAG, $p=2$) \\
{\normalsize \ding{195}} &\textcolor{cyan}{\CheckmarkBold} &\textcolor{cyan}{\CheckmarkBold} &-          &-           &4.67 (\textsf{\footnotesize S3}, $lr$=1e-3, same  $\beta$, w/o NAG, $p=2$) \\
{\normalsize \ding{196}} &\textcolor{cyan}{\CheckmarkBold} &\textcolor{cyan}{\CheckmarkBold} &\textcolor{cyan}{\CheckmarkBold} &-           &4.64 (\textsf{\footnotesize S3}, $lr$=1e-3, same  $\beta$, w/ NAG,  $p=2$) \\
{\normalsize \ding{197}} &\textcolor{cyan}{\CheckmarkBold} &\textcolor{cyan}{\CheckmarkBold} &\textcolor{cyan}{\CheckmarkBold} &\textcolor{cyan}{\CheckmarkBold}  &4.60 (\textsf{\footnotesize S3}, $lr$=3e-3, same  $\beta$, w/ NAG,  $p=3$) \\
\hline
\end{tabular}
\label{Tab.4}
\end{table*}

 \begin{figure*}[!th]
 \centering
 \vspace{-0pt}
  \subfloat{\includegraphics[width=.65\textwidth]{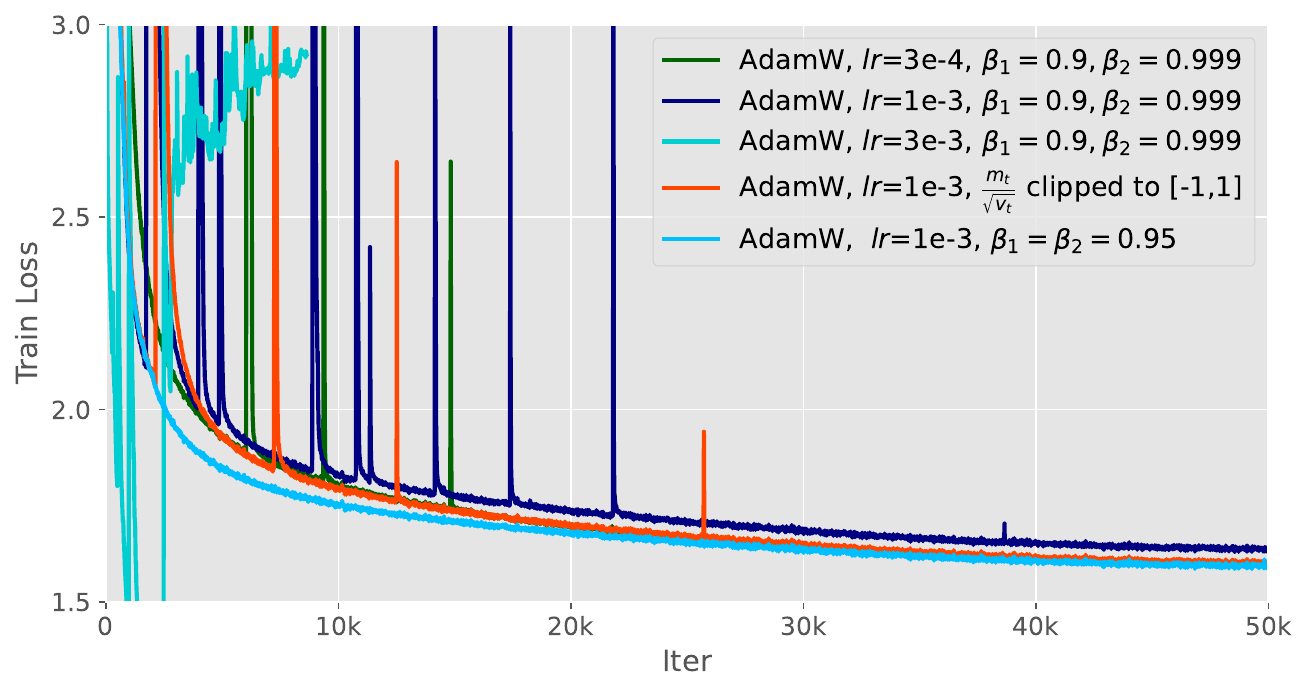}}
    \vspace{-10pt}
  \caption{\small{ The Loss spikes phenomenons during Training GPT-2 (345M) on OpenWebText using AdamW.} }
  \label{Fig.8}
\end{figure*}

We also implement ablation experiments for training GPT-2(345M).  Figure \ref{Fig.6_5} and Table \ref{Tab.4} showcase that employing a large learning rate and sharing the same $\beta$ alone have little or even a negative impact on performance (e.g., Exp. \ding{172} vs. Exp. \ding{173}, Exp. \ding{172} vs. Exp. \ding{174}), while their combination results in a notable improvement (e.g., Exp. \ding{173}, Exp. \ding{174} vs. Exp. \ding{175}). In contrast,  the performance gain from the benefits of NAG is more pronounced (e.g., Exp. \ding{175} vs. Exp. \ding{176}), and harnessing a larger $p$ can also have an individually beneficial effect on performance (e.g., Exp. \ding{176} vs. Exp. \ding{177}),

% \begin{figure*}[h]
% \centering
% \vspace{-0pt}
%  \subfloat{\includegraphics[width=.7\textwidth]{./figures/gpt2_350m_ablation_convergence.pdf}}
%    \vspace{-10pt}
%  \caption{\small{ Ablation study on train loss of \textsf{\footnotesize S3} for training GPT-2(345M)  on OpenWebText.} }
%  \label{Fig.10}
%  \vspace{-5pt}
%\end{figure*}

\subsection{Sensitivity Analysis for Hyperparameters \label{Sensitivity Analysis}}
We perform a grid search to verify the sensitivity to the momentum order $p$ and the momentum coefficient $\beta$ of \textsf{\small S3} on ViT-B/16 with $150$ training epoches. As shown in Figure \ref{Fig.7}, all combinations achieve an accuracy of $80.20\%+$, surpassing the $80.13\%$ accuracy achieved by \textsf{\small Adam} with $300$ training epochs. The performance of \textsf{\small S3} is not sensitive when $p>1$, and $p=1$ achieves a slightly lower accuracy. However, as pointed out in Section 3, the computation cost becomes  lower when $p=1$. Another interesting phenomenon is that setting $\beta$ to $0.95$ obtains the highest accuracies across different $p$, and $p=3$ performs well in most cases.

\subsection{Verifying the Reason for Loss Spikes from Adam}

In this subsection, we further experimentally verify the claim that the potential for excessively large updates in \textsf{\small Adam} with relatively large learning rates is the underlying cause of loss spikes, as discussed in Section 3. Figure \ref{Fig.8} visually illustrates that convergence with vanilla \textsf{\small Adam} is achieved at the baseline learning rate of $3 \times 10^{-4}$, despite occasional spikes. However, when the learning rate is increased to $1 \times 10^{-3}$, more frequent spikes and higher loss values are observed, even with the same number of iterations. Furthermore, employing a $10 \times$ higher learning rate of $3 \times 10^{-3}$ leads to premature divergence with pronounced spikes. All these phenomena are consistent with the analysis in Section 2.

As shown in Figure \ref{Fig.8}, naively clipping the \textsf{\small Adam} updates to the range $[-1, 1]$ reduces the frequency of loss spikes but does not completely eliminate them. This suggests that fine-tuning the clipped value is necessary to balance performance, which adds complexity due to the need to tune an additional hyperparameter. In contrast, as proven in Theorem 2 in Section 4, minimizing the maximum update of \textsf{\small Adam} by setting $\beta_1 = \beta_2$ completely eliminates loss spikes. This further confirms the validity of our analysis in Sections 3 and 4.

\section{Conclusion and Discussion}

In this paper, we thoroughly examine the strengths and weaknesses of the widely used optimizer \textsf{\small Adam}. Building on our analysis, we propose \textsf{\small S3}, an innovative optimizer that incorporates three key improvements over \textsf{\small Adam}. Comprehensive experiments across vision and language tasks demonstrate \textsf{\small S3}'s enhanced training efficiency and superior inference performance. Furthermore, we challenge the conventional belief that \textsf{\small Adam}'s effectiveness is due to simplifying second-order descent, instead showing that its success is rooted in sign-like descent. This insight lays the foundation for the development of more advanced optimizers. Additionally, we present the first theoretical proof of adaptive optimizer convergence from the perspective of sign descent. Most notably, we identify the root cause of loss spikes during LLM training and propose an effective solution, offering substantial benefits for the community in the LLM era.

\bibliography{./refs/refs}
\bibliographystyle{IEEEtran}

\newpage
\onecolumn

\appendix
\section{Theoretical Proofs}

\subsection {Proof of Theorem 1}

\textbf{Proof.} Recalling Eq. (\ref{Eq.Adam}), we know
\begin{equation}
\begin{aligned}
& \bm{m}_t^{(j)} = \frac{1-\beta_1}{1-\beta_1^t}\sum_{k=1}^t \beta_1^{t-k} \bm{g}_k^{(j)} \\
& \bm{v}_t^{(j)} = \frac{1-\beta_2}{1-\beta_2^t}\sum_{k=1}^t \beta_2^{t-k} (\bm{g}_k^{(j)})^2.
\end{aligned}
\end{equation}
Then,
\begin{equation}
\begin{aligned}
\frac{\vert\bm{m}_t^{(j}\vert}{\sqrt{\bm{v}_t^{(j)}}} =& \frac{(1-\beta_1)\sqrt{1-\beta_2^t}}{(1-\beta_1^t)\sqrt{1-\beta_2}}\cdot \frac{\vert\sum_{k=1}^t \beta_1^{t-k} \bm{g}_k^{(j)}\vert}{\sqrt{\sum_{k=1}^t \beta_2^{t-k} (\bm{g}_k^{(j)})^2}} \\
\overset{(i)}{\le} & \frac{(1-\beta_1)\sqrt{1-\beta_2^t}}{(1-\beta_1^t)\sqrt{1-\beta_2}}\cdot \frac{\sum_{k=1}^t \beta_1^{t-k} \vert\bm{g}_k^{(j)}\vert}{\sqrt{\sum_{k=1}^t \beta_2^{t-k} (\bm{g}_k^{(j)})^2}} \\
\overset{(ii)}{\le}& \frac{(1-\beta_1)\sqrt{1-\beta_2^t}}{(1-\beta_1^t)\sqrt{1-\beta_2}}\cdot \frac{\sqrt{\sum_{k=1}^t \beta_2^{t-k} (\bm{g}_k^{(j)})^2}\sqrt{\sum_{k=1}^t\frac{\beta_1^{2(t-k)}}{\beta_2^{t-k}}}}{\sqrt{\sum_{k=1}^t \beta_2^{t-k} (\bm{g}_k^{(j)})^2}} \\
=& \frac{(1-\beta_1)\sqrt{1-\beta_2^t}}{(1-\beta_1^t)\sqrt{1-\beta_2}}\cdot  \sqrt{\sum_{k=1}^t\frac{\beta_1^{2(t-k)}}{\beta_2^{t-k}}} \\
\overset{(iii)}{=}& \frac{(1-\beta_1)\sqrt{1-\beta_2^t}\sqrt{1-(\frac{\beta_1^2}{\beta_2})^t}}{(1-\beta_1^t)\sqrt{1-\beta_2}\sqrt{1-\frac{\beta_1^2}{\beta_2}}} \\
\simeq & \frac{1-\beta_1}{\sqrt{1-\beta_2}\sqrt{1-\frac{\beta_1^2}{\beta_2}}},
%\le & \frac{(1-\beta_1)}{(1-\beta_1^t)\sqrt{1-\beta_2}\sqrt{1-\frac{\beta_1^2}{\beta_2}}},
\end{aligned}
\end{equation}

where $(i)$ holds due to the fact $\vert\bm{a}^{(j)} + \bm{b}^{(j)}\vert \le \vert\bm{a}^{(j)}\vert + \vert\bm{b}^{(j)}\vert $; $(ii)$ holds due to Cauchy-Schiwaz inequality; $(iii)$ holds due to $\beta_1^2 \le \beta_2$. $\frac{\vert\bm{m}_t^{(j}\vert}{\sqrt{\bm{v}_t^{(j)}}}$  reach to the largest value if the signs of $\{\bm{g}_t^{(j)}, \bm{g}_{t-1}^{(j)}, ...\}$ are the same and $\vert\bm{g}_t^{(j)}\vert = \frac{\beta_2\vert\bm{g}_{t-1}^{(j)}\vert}{\beta_1} = \frac{\beta_2^2\vert\bm{g}_{t-2}^{(j)}\vert}{\beta_1^2} = ...=\frac{\beta_2^k\vert\bm{g}_{t-k}^{(j)}\vert}{\beta_1^k}...$.

\subsection {Proof of Theorem 2}

\textbf{Proof.} (1). According to \textsf{\small S3} in Algorithm 1, we have
\begin{equation}
\begin{aligned}
&\bm{n}_t^{(j)} = (1-\beta_1)\left(\sum_{k=1}^t \beta_1^{t-k+1} \bm{g}_k^{(j)} + \bm{g}_t^{(j)}\right) \\
&\bm{b}_t^{(j)}(p) = \left((1-\beta_2)\left(\sum_{k=1}^t \beta_2^{t-k+1}
\vert\bm{g}_k^{(j)}\vert^p + \vert\bm{g}_t^{(j)}\vert^p\right)\right)^{\sfrac{1}{p}}
\end{aligned}
\end{equation}

Then, assuming $q$ satisfies  $\frac{1}{p}+\frac{1}{q} =1 $, we obtain
\begin{equation}
\begin{aligned}
\frac{\vert\bm{n}_t^{(j)}\vert}{\bm{b}_t^{(j)}(p)} = & \frac{1-\beta_1}{(1-\beta_2)^{\sfrac{1}{p}}}\cdot \frac{\vert\sum_{k=1}^t \beta_1^{t-k+1} \bm{g}_k^{(j)} + \bm{g}_t^{(j)}\vert}{\left(\sum_{k=1}^t \beta_2^{t-k+1} \vert\bm{g}_k^{(j)}\vert^p + \vert\bm{g}_t^{(j)}\vert^p\right)^{\sfrac{1}{p}}} \\
\overset{(i)}{\le} & \frac{1-\beta_1}{(1-\beta_2)^{\sfrac{1}{p}}}\cdot \frac{\sum_{k=1}^t \beta_1^{t-k+1} \vert\bm{g}_k^{(j)}\vert + \vert\bm{g}_t^{(j)}\vert}{\left(\sum_{k=1}^t \beta_2^{t-k+1} \vert\bm{g}_k^{(j)}\vert^p + \vert\bm{g}_t^{(j)}\vert^p\right)^{\sfrac{1}{p}}} \\
\overset{(ii)}{\le} &  \frac{1-\beta_1}{(1-\beta_2)^{\sfrac{1}{p}}} \cdot \frac{\left(\sum_{k=1}^t \beta_2^{t-k+1} \vert\bm{g}_k^{(j)}\vert^p + \vert\bm{g}_t^{(j)}\vert^p\right)^{\sfrac{1}{p}}\left(\sum_{k=1}^t\left(\frac{\beta_1^{(t-k+1)}}{\beta_2^{\frac{1}{p}(t-k+1)}}\right)^q + 1\right)^{\sfrac{1}{q}}}{\left(\sum_{k=1}^t \beta_2^{t-k+1} \vert\bm{g}_k^{(j)}\vert^p + \vert\bm{g}_t^{(j)}\vert^p\right)^{\sfrac{1}{p}}} \\
=& \frac{1-\beta_1}{(1-\beta_2)^{\sfrac{1}{p}}} \cdot \left(\sum_{k=1}^t\left(\frac{\beta_1^{(t-k+1)}}{\beta_2^{\frac{1}{p}(t-k+1)}}\right)^q + 1\right)^{\sfrac{1}{q}} \\
\overset{(iii)}{\le} & \frac{1-\beta_1}{(1-\beta_2)^{\sfrac{1}{p}}\left(1 - \frac{\beta_1^q}{\beta_2^{\sfrac{q}{p}}}\right)^{\sfrac{1}{q}}},
\end{aligned}
\end{equation}
where  $(i)$ holds due to the fact $\vert a + b\vert \le \vert a\vert + \vert b\vert $; $(ii)$ holds due to H\"{o}lder inequality $\sum_{i=1}^s a_i b_i \le (\sum_{i=1}^s a_i^p)^{\sfrac{1}{p}}(\sum_{i=1}^s b_i^q)^{\sfrac{1}{q}}$ if $\frac{1}{p}+\frac{1}{q} =1 $ and $p, q \ge 1$; $(iii)$ holds due to $\beta_1 \le \beta_2^{\sfrac{1}{p}}$.

(2). The upper bound of  each element of the update $\frac{\bm{n}_t^{(j)}}{\bm{b}_t^{(j)}}$ can be
\begin{equation}
\begin{aligned}
\frac{1-\beta_1}{(1-\beta_2)^{\sfrac{1}{p}}\left(1 - \frac{\beta_1^q}{\beta_2^{\sfrac{q}{p}}}\right)^{\sfrac{1}{q}}}
\overset{(i)}{\ge} & \frac{1-\beta_1}{\frac{1}{p}(1-\beta_2)+\frac{1}{q}(1 - \frac{\beta_1^q}{\beta_2^{\sfrac{q}{p}}})} \\
\overset{(ii)}{=} & \frac{1-\beta_1}{1-(\frac{\beta_2}{p}+ \frac{\beta_1^q}{q\beta_2^{\sfrac{q}{p}}})} \\
\overset{(iii)}{\ge} & \frac{1-\beta_1}{1-\beta_2^{\sfrac{1}{p}}\frac{\beta_1}{\beta_2^{\sfrac{1}{p}}}} \\
=& \frac{1-\beta_1}{1-\beta_1} \\
=& 1,
\end{aligned}
\end{equation}
where$(i)$ holds due to  Young's inequality $\frac{a}{p} + \frac{b}{q} \ge a^{\sfrac{1}{p}}b^{\sfrac{1}{q}}$; $(ii)$ holds owing to  $\frac{1}{p}+\frac{1}{q} =1 $; $(iii)$ holds resulting from  Young's inequality again.

Notably,  according to the property of  Young's inequality, the equality in $(i)$ and $(iii)$ can be reached, if and only if
\begin{equation}
\beta_2 = \frac{\beta_1^q}{\beta_2^{\sfrac{q}{p}}} \quad \Rightarrow \quad \beta_1^q = \beta_2^{1+\frac{q}{p}} \quad \Rightarrow \quad \beta_1^q = \beta_2^{1+(1-\frac{1}{q})q} \quad \Rightarrow \quad \beta_1^q = \beta_2^q \quad \Rightarrow \quad \beta_1 = \beta_2
\end{equation}

Therefore, when $\beta_1=\beta_2$, the  upper bound of  each element of the update $\frac{\bm{n}_t^{(j)}}{\bm{b}_t^{(j)}}$ reaches to the smallest $1$.

(3). Following \textsf{\small S3} in Algorithm 1, we have
\begin{equation}
\begin{aligned}
& \bm{b}_t^{(j)}(p_1) = \left((1-\beta)\left(\sum_{k=1}^t \beta^{t-k+1}
\vert\bm{g}_k^{(j)}\vert^{p_1} + \vert\bm{g}_t^{(j)}\vert^{p_1}\right)\right)^{\sfrac{1}{p_1}}\\
& \bm{b}_t^{(j)}(p_2) = \left((1-\beta)\left(\sum_{k=1}^t \beta^{t-k+1}
\vert\bm{g}_k^{(j)}\vert^{p_2} + \vert\bm{g}_t^{(j)}\vert^{p_2}\right)\right)^{\sfrac{1}{p_2}}.
\end{aligned}
\end{equation}

Denoting $r=\frac{p_2}{p_1}$, we then obtain
\begin{equation}
\begin{aligned}
(\bm{b}_t^{(j)}(p_1))^{p_2} = (\bm{b}_t^{(j)}(p_1))^{rp_1} =& \left((1-\beta)\left(\sum_{k=1}^t \beta^{t-k+1}
\vert\bm{g}_k^{(j)}\vert^{p_1} + \vert\bm{g}_t^{(j)}\vert^{p_1}\right)\right)^r \\
\le &  (1-\beta)\left(\sum_{k=1}^t \beta^{t-k+1}
\vert\bm{g}_k^{(j)}\vert^{rp_1} + \vert\bm{g}_t^{(j)}\vert^{rp_1}\right) \\
=& (1-\beta)\left(\sum_{k=1}^t \beta^{t-k+1}
\vert\bm{g}_k^{(j)}\vert^{p_2} + \vert\bm{g}_t^{(j)}\vert^{p_2}\right)\\
=&(\bm{b}_t^{(j)}(p_2))^{p_2},
\end{aligned}
\end{equation}
where the inequality holds due to Jensen's inequality and the fact $(1-\beta)(\sum_{k=1}^t \beta^{t-k+1} +1) < 1$.

\subsection{Proof of Theorem 3}

\textbf{Proof.} We first deduce from ASGD(I) to ASGD(II). According to (I), we have
\begin{equation}
\begin{aligned}
\tilde{ \bm{x}}_{t+1} =& \tilde{ \bm{x}}_t - \gamma  \bm{m}_t \\
=& \tilde{ \bm{x}}_t - \gamma (\beta  \bm{m}_{t-1} + (1-\beta)\nabla f(\tilde{ \bm{x}}_t-\gamma\beta  \bm{m}_{t-1};\zeta_t ))
\end{aligned}
\end{equation}

Subtracting $\gamma \beta  \bm{m}_{t}$ on both sides, we obtain
\begin{equation}
\tilde{ \bm{x}}_{t+1} - \gamma \beta  \bm{m}_{t} = \tilde{ {x}}_t - \gamma \beta  \bm{m}_{t-1} - \gamma(\beta \bm{m}_{t} - (1-\beta)\nabla f(\tilde{ \bm{x}}_t-\gamma\beta  \bm{m}_{t-1};\zeta_t ))
\end{equation}

Setting $ \bm{x}_t=\tilde{ \bm{x}}_t-\gamma\beta \bm{m}_{t-1}$, we further have
\begin{equation}
{ \bm{x}}_{t+1} = { \bm{x}}_t - \gamma (\beta \bm{m}_t + (1-\beta) \nabla f( \bm{x}_t;\zeta_t ))
\end{equation}

Thus, ASGD(I) becomes ASGD(II).

Then, we deduce from  ASGD(III). Denoting $ \bm{n}_t =  \beta  \bm{m}_t + (1-\beta) \bm{g}_t$, we have
\begin{equation}
\begin{aligned}
 \bm{n}_{t} - \beta \bm{n}_{t-1} =&  \beta  \bm{m}_t + (1-\beta) \bm{g}_t -  \beta \bm{n}_{t-1} \\
= & (1-\beta) \bm{g}_t + \beta (\beta \bm{m}_{t-1}+ (1-\beta) \bm{g}_t) -  \beta \bm{n}_{t-1} \\
= & (1-\beta) \bm{g}_t + \beta (\beta \bm{m}_{t-1}+ (1-\beta) \bm{g}_t) -  \beta(\beta \bm{m}_{t-t}+ (1-\beta) \bm{g}_{t-1}) \\
= & (1-\beta) \bm{g}_t + \beta (1-\beta) ( \bm{g}_t -  \bm{g}_{t-1}) \\
= & (1-\beta) ( \bm{g}_t + \beta(\bm{g}_t -  \bm{g}_{t-1})).
\end{aligned}
\end{equation}

It indicates $ \bm{n}_t =  \bm{m}_t + \beta  \bm{r}_t$ where $ \bm{m}_t = \beta  \bm{m}_{t-1} + (1-\beta) \bm{g}_t$ and  $ \bm{r}_t = \beta  \bm{r}_{t-1} + (1-\beta)( \bm{g}_t -  \bm{g}_{t-1})$. Therefore, ASGD (II) is equivalent to  ASGD (III).

\subsection{Auxiliary Lemma }

\begin{lemma}
Under Assumption 2, for any $\bm{x}, \bm{y} \in {\mathbb R}^d $ with $\Vert \bm{x} - \bm{y}\Vert_2 \le R$, the function obeys

\begin{equation}
F(\bm{y}) \le  F(\bm{x}) + \langle \nabla F(\bm{x}), \bm{y} -\bm{x} \rangle + \frac{L_0 + L_1\Vert \nabla F(\bm{x}) \Vert_2}{2}\Vert \bm{y} - \bm{x}\Vert_2^2.
\end{equation}
\label{lamma_smooth}
\end{lemma}
\textbf{Proof.}  For any $\bm{x}, \bm{y} \in {\mathbb R}^d $ with $\Vert \bm{x} - \bm{y}\Vert_2 \le R$, we have
\begin{equation}
\begin{aligned}
F(\bm{y}) = & F(\bm{x}) + \int_{0}^1 \langle \nabla F(\bm{x} + t(\bm{y}-\bm{x})), \bm{y}-\bm{x} \rangle dt \\
= & F(\bm{x}) + \langle \nabla F(\bm{x}), \bm{y}-\bm{x} \rangle + \int_{0}^1 \langle \nabla F(\bm{x} + t(\bm{y}-\bm{x}))-\nabla F(\bm{x}), \bm{y}-\bm{x} \rangle dt \\
\overset{(i)}{\le} & F(\bm{x}) + \langle \nabla F(\bm{x}), \bm{y}-\bm{x} \rangle + \int_0^1 \Vert \nabla F(\bm{x} + t(\bm{y}-\bm{x}))-\nabla F(\bm{x})\Vert_2 \Vert\bm{y}-\bm{x}\Vert_2 dt \\
\overset{(ii)}{\le} & F(\bm{x}) + \langle \nabla F(\bm{x}), \bm{y}-\bm{x} \rangle + (L_0 + L_1 \nabla F(\bm{x}))\Vert\bm{y}-\bm{x}\Vert_2^2 \int_0^1 t d_t \\
=&  F(\bm{x}) + \langle \nabla F(\bm{x}), \bm{y} -\bm{x} \rangle + \frac{L_0 + L_1\Vert \nabla F(\bm{x}) \Vert}{2}\Vert \bm{y} - \bm{x}\Vert_2^2,
\end{aligned}
\end{equation}
where$(i)$ holds due to Cauchy-Schwarz inequality, and  $(ii)$ holds due to Assumption 2.

\subsection{Proof of Theorem 4}

\textbf{Proof.} Following Lemma \ref{lamma_smooth} with $\bm{x}_{t+1}\rightarrow \bm{y}$ and $\bm{x}_{t}\rightarrow \bm{x}$, we have
\begin{equation}
F(\bm{x}_{t+1}) \le  F(\bm{x}_t) + \langle \nabla F(\bm{x}_t), \bm{x}_{t+1} -\bm{x}_t \rangle + \frac{L_0 + L_1\Vert \nabla F(\bm{x}_t) \Vert_2}{2}\Vert \bm{x}_{t+1} - \bm{x}_t\Vert_2^2.
\end{equation}
Recalling the update rule $\bm{x}_{t+1} = \bm{x}_t - \gamma \frac{\bm{n}_t}{\bm{b}_t} = \bm{x}_t - \gamma \frac{\vert \bm{n}_t \vert}{\bm{b}_t} \circ \frac{\bm{n}_t}{|\bm{n}_t|} = \bm{x}_t - \gamma \bm{u}_t \circ {\rm Sign}(\bm{n}_t) $, we further obtain
\begin{equation}
\begin{aligned}
F(\bm{x}_{t+1}) \le &  F(\bm{x}_t) - \langle \nabla F(\bm{x}_t), \gamma \bm{u}_t \circ {\rm Sign}(\bm{n}_t) \rangle + \frac{(L_0 + L_1\Vert \nabla F(\bm{x}_t) \Vert_2)\gamma^2}{2}\Vert \bm{u}_t \Vert_2^2 \\
= & F(\bm{x}_t) - \langle \nabla F(\bm{x}_t), \gamma \bm{u}_t \circ {\rm Sign}(\nabla F(\bm{x}_t)) \rangle + \underbrace{ \left\langle \nabla F(\bm{x}_t), \gamma \bm{u}_t \circ ({\rm Sign}(\nabla F(\bm{x}_t)) - {\rm Sign}(\bm{n}_t) )\right\rangle}_{T_1}  \\
&+\frac{(L_0 + L_1\Vert \nabla F(\bm{x}_t) \Vert)\gamma^2}{2}\Vert \bm{u}_t \Vert_2^2.
\end{aligned}
\label{Eq.s_7}
\end{equation}

There are two cases for each element of $T_1$. If ${\rm Sign}(\nabla F(\bm{x}_t)^{(j)}) = {\rm Sign}(\bm{n}_t^{(j)})$, $  \nabla F(\bm{x}_t)^{(j)} \cdot \bm{u}_t^{(j)} \cdot\left ({\rm Sign}(\nabla F(\bm{x}_t))^{(j)} - {\rm Sign}(\bm{n}_t^{(j)}) \right)=0$. If ${\rm Sign}(\nabla F(\bm{x}_t)^{(j)}) \neq {\rm Sign}(\bm{n}_t^{(j)})$,  $ \nabla F(\bm{x}_t)^{(j)} \cdot \bm{u}_t^{(j)} \cdot\left ({\rm Sign}(\nabla F(\bm{x}_t))^{(j)} - {\rm Sign}(\bm{n}_t^{(j)}) \right)=2\bm{u}_t^{(j)}\vert\nabla F(\bm{x}_t)^{(j)}\vert \le 2\bm{u}_t^{(j)}\vert\nabla F(\bm{x}_t)^{(j)}- \bm{n}_t^{(j)}\vert$, hence $T_1 = 2\sum_{j=1}^d\bm{u}_t^{(j)}\vert\nabla F(\bm{x}_t)^{(j)}- \bm{n}_t^{(j)}\vert$.

Rearranging Eq. (\ref{Eq.s_7}), we have

\begin{equation}
\begin{aligned}
F(\bm{x}_{t+1}) \le & F(\bm{x}_t) - \left\langle \nabla F(\bm{x}_t), \gamma \bm{u}_t \circ {\rm Sign}(\nabla F(\bm{x}_t)) \right\rangle +2\gamma\sum_{j=1}^d\bm{u}_t^{(j)}\vert\nabla F(\bm{x}_t)^{(j)}- \bm{n}_t^{(j)}\vert  \\
&+\frac{(L_0 + L_1\Vert \nabla F(\bm{x}_t) \Vert_2)\gamma^2}{2}\Vert \bm{u}_t \Vert_2^2 \\
\le&  F(\bm{x}_t) - \gamma u_{\min} \Vert\nabla F(\bm{x}_t)\Vert_1  + 2\gamma \Vert \bm{n}_t - \nabla F(\bm{x}_t)\Vert_1+\frac{\gamma^2d(L_0 + L_1\Vert \nabla F(\bm{x}_t) \Vert_2)}{2} \\
\le&  F(\bm{x}_t) - \gamma u_{\min} \Vert\nabla F(\bm{x}_t)\Vert_1  + 2\gamma\sqrt{d} \Vert \bm{n}_t - \nabla F(\bm{x}_t)\Vert_2+\frac{\gamma^2d(L_0 + L_1\Vert \nabla F(\bm{x}_t) \Vert_1)}{2}
\end{aligned}
\label{Eq.s_8}
\end{equation}

where the second inequality holds due to $0<u_{\min} \le \bm{u}_t^{(j)} \le 1$, and the third inequality holds owing to the fact $\Vert \bm{a}\Vert_2 \le \Vert \bm{a}\Vert_1 \le \sqrt{d}\Vert \bm{a}\Vert_2$ for any $\bm{a} \in {\mathbb R}^d$.

Summing over $1$ to $T$ and taking expectation on it, we have
\begin{equation}
\frac{(u_{\min}-\frac{\gamma dL_1}{2})}{T}\sum_{t=1}^T \mathbb E [\Vert\nabla F(\bm{x}_t)\Vert_1] \le \frac{ F(\bm{x}_{1}) - F(\bm{x}^*)}{\gamma T} + \frac{2\sqrt{d}}{T}\sum_{t=1}^T {\mathbb E} [\Vert \bm{n}_t - \nabla F(\bm{x}_t)\Vert_2] + \frac{\gamma d L_0}{2}
\label{Eq.s_9}
\end{equation}

Recalling $\bm{m}_t = \beta\bm{m}_{t-1} + (1-\beta)\bm{g}_t$, we obtain

\begin{equation}
\begin{aligned}
 \bm{m}_t - \nabla F(\bm{x}_t) =&  \left(\beta\bm{m}_{t-1} + (1-\beta)\bm{g}_t\right) - \nabla F(\bm{x}_t) \\
=& \beta(\bm{m}_{t-1} - \nabla F(\bm{x}_{t-1})) +(1-\beta)(\bm{g}_t -  \nabla F(\bm{x}_{t}))
 - \beta(\nabla F(\bm{x}_t) - \nabla F(\bm{x}_{t-1})).
\end{aligned}
\end{equation}

%Under Assumption 2, we obtain
%\begin{equation}
%\begin{aligned}
% \Vert \bm{n}_t - \nabla F(\bm{x}_t)\Vert_2 \le &  \beta \Vert \bm{n}_{t-1} - \nabla F(\bm{x}_{t-1})\Vert_2 + (1-\beta^2)\Vert\bm{g}_t -  \nabla F(\bm{x}_t)\Vert_2 \\
%  & +(1-\beta)\beta\Vert\bm{g}_{t-1} -  \nabla F(\bm{x}_{t-1})\Vert_2 + \beta^2 \Vert \nabla F(\bm{x}_t) - \nabla F(\bm{x}_{t-1})\Vert_2 \\
%  \le & \beta \Vert \bm{n}_{t-1} - \nabla F(\bm{x}_{t-1})\Vert_2  + (1-\beta^2)\Vert\bm{g}_t -  \nabla F(\bm{x}_t)\Vert_2 \\
%  & +(1-\beta)\beta\Vert\bm{g}_{t-1} -  \nabla F(\bm{x}_{t-1})\Vert_2 + {\gamma \beta^2 \sqrt{d} (L_0 + L_1\Vert \nabla F(\bm{x}_t) \Vert_2)}
% \end{aligned}
%\end{equation}

Utilizing recursion, we further have
\begin{equation}
\begin{aligned}
 \bm{m}_t - \nabla F(\bm{x}_t) =& -\beta^t\nabla F(\bm{x}_{1})+ (1-\beta)\sum_{k=1}^t \beta^{t-k} (\bm{g}_k -  \nabla F(\bm{x}_k))
  - \sum_{k=1}^t \beta^{t-k+1}(\nabla F(\bm{x}_k) - \nabla F(\bm{x}_{k-1})),
\end{aligned}
\end{equation}
where $\bm{m}_1 - \nabla F(\bm{x}_{1}) = -\beta_1 \nabla F(\bm{x}_{1})+(1-\beta_1)(\bm{g}_1-\nabla F(\bm{x}_{1}))$ due to $\bm{m}_0=0$.

Hence,
\begin{equation}
\begin{aligned}
 \bm{n}_t - \nabla F(\bm{x}_t) =& \beta (\bm{m}_t - \nabla F(\bm{x}_t)) + (1-\beta)(\bm{g}_t - \nabla F(\bm{x}_t)) \\
 =& -\beta^{t+1}\nabla F(\bm{x}_{1})+ (1-\beta)\left(\sum_{k=1}^t \beta^{t-k+1} (\bm{g}_k -  \nabla F(\bm{x}_k)+ (\bm{g}_t - \nabla F(\bm{x}_t))\right)\\
  &- \beta^2\sum_{k=1}^t \beta^{t-k}(\nabla F(\bm{x}_k) - \nabla F(\bm{x}_{k-1})),
 \end{aligned}
\end{equation}

Then, we obtain

\begin{equation}
\begin{aligned}
\frac{1}{T}\sum_{t=1}^T {\mathbb E}\left[\Vert \bm{n}_t - \nabla F(\bm{x}_t) \Vert_2\right]  \le & \underbrace{\frac{\beta}{T}\sum_{t=1}^T \beta^{t}\left\Vert \nabla F(\bm{x}_1)\right\Vert_2}_{T_2} \\
&+ \underbrace{\frac{1-\beta}{T}\sum_{t=1}^T {\mathbb E}\left[\left\Vert\sum_{k=1}^t \beta^{t-k+1} (\bm{g}_k -  \nabla F(\bm{x}_k))\right\Vert_2  + \Vert \bm{g}_t - \nabla F(\bm{x}_t) \Vert_2\right]}_{T_3} \\
 &+  \underbrace{\frac{\beta^2}{T}\sum_{t=1}^T {\mathbb E}\left[\left\Vert \sum_{k=1}^t \beta^{t-k}(\nabla F(\bm{x}_k) - \nabla F(\bm{x}_{k-1}))\right\Vert_2 \right]}_{T_4} \\
\end{aligned}
\label{Eq.s_11}
\end{equation}

In terms of $T_2$, we obtain
\begin{equation}
T_2 = \frac{\beta}{T}\sum_{t=1}^T \beta^{t}\left\Vert \nabla F(\bm{x}_1)\right\Vert_2 \le \frac{\beta}{(1-\beta)T}\left\Vert \nabla F(\bm{x}_1)\right\Vert_2
\end{equation}

As for $T_3$, we have
\begin{equation}
\begin{aligned}
T_3 =&\frac{1-\beta}{T}\sum_{t=1}^T {\mathbb E}\left[\left\Vert\sum_{k=1}^t \beta^{t-k+1} (\bm{g}_k -  \nabla F(\bm{x}_k))\right\Vert_2 + \Vert \bm{g}_t - \nabla F(\bm{x}_t) \Vert_2 \right] \\
\overset{(i)}{\le} & \frac{1-\beta}{T}\sum_{t=1}^T\sqrt{{\mathbb E}\left[\left\Vert\sum_{k=1}^t \beta^{t-k+1} (\bm{g}_k -  \nabla F(\bm{x}_k))\right\Vert_2^2 + \Vert \bm{g}_t - \nabla F(\bm{x}_t) \Vert_2^2 \right]} \\
\overset{(ii)}{=} & \frac{1-\beta}{T}\sum_{t=1}^T\sqrt{\sum_{k=1}^t \beta^{2(t-k+1)} {\mathbb E}\left[\left\Vert\bm{g}_k -  \nabla F(\bm{x}_k)\right\Vert_2^2 + \Vert \bm{g}_t - \nabla F(\bm{x}_t) \Vert_2 \right]} \\
\overset{(iii)}{\le} &\frac{1-\beta}{T}\sum_{t=1}^T \sqrt{\sum_{k=1}^{t+1} \beta^{2(t-k+1)}\sigma^2} \\
\le & \frac{1-\beta}{\sqrt{1-\beta^2}}\sigma \\
\le & \sqrt{1-\beta}\sigma,
\end{aligned}
\end{equation}
where $(i)$ holds due to the fact $({\mathbb E}[Z])^2 \le {\mathbb E}[Z^2] $; $(ii)$ holds owing to ${\mathbb E}[\bm{g}_k -  \nabla F(\bm{x}_k)]=\bm{0} $ according to Assumption 3; $(iii)$ holds resulting from ${\mathbb E}\left[\left\Vert\bm{g}_k -  \nabla F(\bm{x}_k)\right\Vert_2^2\right]\le \sigma^2$ according to Assumption 3.

Now we turn attention to $T_4$, \emph{i.e.},
\begin{equation}
\begin{aligned}
T_4 = & \frac{\beta^2}{T}\sum_{t=1}^T{\mathbb E}\left[\left\Vert  \beta^{t-k}(\nabla F(\bm{x}_k) - \nabla F(\bm{x}_{k-1}))\right\Vert_2 \right]  \\
\overset{(i)}{\le} & \frac{\beta^2}{T}\sum_{t=1}^T\sum_{k=1}^t\beta^{t-k}{\mathbb E}\left[\left\Vert \nabla F(\bm{x}_k) - \nabla F(\bm{x}_{k-1}) \right\Vert_2 \right] \\
\overset{(ii)}{\le} & \frac{\beta^2}{T}\sum_{t=1}^T \sum_{k=1}^t\beta^{t-k}{\mathbb E}\left[(L_0 + L_1\Vert \nabla F(\bm{x}_{k}) \Vert_2)\Vert\bm{x}_k - \bm{x}_{k-1} \Vert_2 \right] \\
\overset{(iii)}{=}&\frac{\beta^2}{T}\sum_{t=1}^T\sum_{k=1}^t\beta^{t-k}{\mathbb E}\left[\gamma(L_0 + L_1\Vert \nabla F(\bm{x}_{k}) \Vert_2)\Vert \bm{u}_{t-1} \Vert_2 \right])\\
\overset{(iv)}{\le} & \frac{\beta^2}{T}\sum_{t=1}^T L_0\gamma\sqrt{d}\sum_{k=1}^t\beta^{t-k} + \beta^2L_1\gamma\sqrt{d}\sum_{k=1}^t\beta^{t-k}\nabla F(\bm{x}_{k}) \\
\le & \frac{\beta^2L_0\gamma\sqrt{d}}{1-\beta} + \beta^2L_1\gamma\sqrt{d}\sum_{t=1}^T\sum_{k=1}^t\beta^{t-k}{\mathbb E}[\Vert\nabla F(\bm{x}_{k})\Vert_2] \\
\overset{(v)}{=}& \frac{\beta^2L_0\gamma\sqrt{d}}{1-\beta} + \frac{\beta^2L_1\gamma\sqrt{d}}{T}\sum_{k=1}^T{\mathbb E}[\Vert\nabla F(\bm{x}_{k})\Vert_2]\sum_{t=k}^T \beta^{t-k} \\
\le & \frac{\beta^2L_0\gamma\sqrt{d}}{1-\beta} + \frac{\beta^2L_1\gamma\sqrt{d}}{(1-\beta)T}\sum_{t=1}^T{\mathbb E}[\Vert\nabla F(\bm{x}_{t})\Vert_2] \\
\overset{(vi)}{\le} & \frac{\beta^2L_0\gamma\sqrt{d}}{1-\beta} + \frac{\beta^2L_1\gamma\sqrt{d}}{(1-\beta)T}\sum_{t=1}^T{\mathbb E}[\Vert\nabla F(\bm{x}_{t})\Vert_1]
\end{aligned}
\label{Eq.s_14}
\end{equation}
where $(i)$ holds due to the fact $\Vert \bm{a} + \bm{b} \Vert_2 \le \Vert \bm{a} \Vert_2 + \Vert\bm{b}\Vert_2$; $(ii)$ holds owing to Assumption 2; $(iii)$ holds due to the update rule; $(iv)$ holds depends on $\bm{u}^{(j)} \le 1$ according to Theorem 2; $(v)$ holds resulting from the fact that $\sum_{i=1}^n\sum_{j=1}^i \bm{a}_{i,j}= \sum_{j=1}^n\sum_{i=j}^n \bm{a}_{i,j}$; $(vi)$ holds due to the fact $\Vert\bm{a}\Vert_2 \le \Vert\bm{a}\Vert_1$.

Combining Eq.(\ref{Eq.s_11}) - Eq.(\ref{Eq.s_14}), we have
\begin{equation}
\begin{aligned}
\frac{1}{T}\sum_{t=1}^{T}{\mathbb E}\left[\Vert \bm{n}_t - \nabla F(\bm{x}_t) \Vert_2\right]  \le & \frac{\beta}{(1-\beta)T} \left\Vert \nabla F(\bm{x}_1)\right\Vert_2 + \sqrt{1-\beta}\sigma \\
& + \frac{\beta^2L_0\gamma\sqrt{d}}{1-\beta} + \frac{\beta^2L_1\gamma\sqrt{d}}{(1-\beta)T}\sum_{t=1}^T{\mathbb E}[\Vert\nabla F(\bm{x}_{t})\Vert_1] \\
\le & \frac{\beta}{(1-\beta)T} \left\Vert \nabla F(\bm{x}_1)\right\Vert_2 +\sqrt{1-\beta}\sigma  \\
 &+ \frac{\beta^2L_0\gamma\sqrt{d}}{1-\beta} + \frac{\beta^2L_1\gamma\sqrt{d}}{(1-\beta)T}\sum_{t=1}^T{\mathbb E}[\Vert\nabla F(\bm{x}_{t})\Vert_1]
\end{aligned}
\label{Eq.s_15}
\end{equation}

Combining  Eq.(\ref{Eq.s_9}) and Eq.(\ref{Eq.s_15}), we obtain
\begin{equation}
\begin{aligned}
\frac{u_{\min}-\frac{\gamma dL_1}{2}-\frac{2\beta^2L_1\gamma\sqrt{d}}{1-\beta}}{T}\sum_{t=1}^T \mathbb E [\Vert\nabla F(\bm{x}_t)\Vert_1] \le& \frac{ F(\bm{x}_{1}) - F(\bm{x}^*)}{\gamma T} + \frac{2\beta\sqrt{d}}{(1-\beta)T}{\mathbb E}\left[\left\Vert \nabla F(\bm{x}_1)\right\Vert_2\right] \\
&+2\sqrt{(1-\beta)d}\sigma +  \frac{2\gamma\beta^2 L_0 {d}}{1-\beta} + \frac{\gamma d L_0}{2}.
\end{aligned}
\label{Eq.s_17}
\end{equation}

Let $\gamma=\frac{1}{L_0T^{\sfrac{3}{4}}}$, $1-\beta = \frac{1}{T^{\sfrac{1}{2}}}$. When $T \ge \max\{(\frac{2dL_1}{L_0u_{\min}})^{\sfrac{4}{3}}, (\frac{8\beta^2\sqrt{d}L_1}{(1-\beta)L_0u_{\min}})^4\}$, we can guarantee
\begin{equation}
u_{\min}-\frac{\gamma dL_1}{2}-\frac{2\gamma\sqrt{d}L_1}{1-\beta} \ge \frac{u_{\min}}{2}.
\end{equation}

Then, setting $U_{\max}=\frac{1}{u_{\min}}$, we reformulate Eq. (\ref{Eq.s_17}) as

\begin{equation}
\begin{aligned}
\frac{1}{T}\sum_{t=1}^T \mathbb E [\Vert\nabla F(\bm{x}_t)\Vert_1] \le& \frac{2L_0U_{\max}(F(\bm{x}_{1})- F(\bm{x}^*))}{T^{\sfrac{1}{4}}} + \frac{4\beta U_{\max}\sqrt{d}{\mathbb E}\left[\left\Vert \nabla F(\bm{x}_1)\right\Vert_2\right]}{T^{\sfrac{1}{2}}} \\
&+\frac{{4}U_{\max}\sqrt{d}\sigma}{T^{\sfrac{1}{4}}} +  \frac{4\beta^2U_{\max} {d}}{T^{\sfrac{1}{4}}} + \frac{U_{\max}d}{T^{\sfrac{7}{4}}}.
\end{aligned}
\label{Eq.s_19}
\end{equation}

\subsection{Detailed Training Setting}

\begin{figure*}[!tb]
  \vspace{-5pt}
 \centering
  \subfloat[\footnotesize{AdamW}]{\includegraphics[width=.42\textwidth]{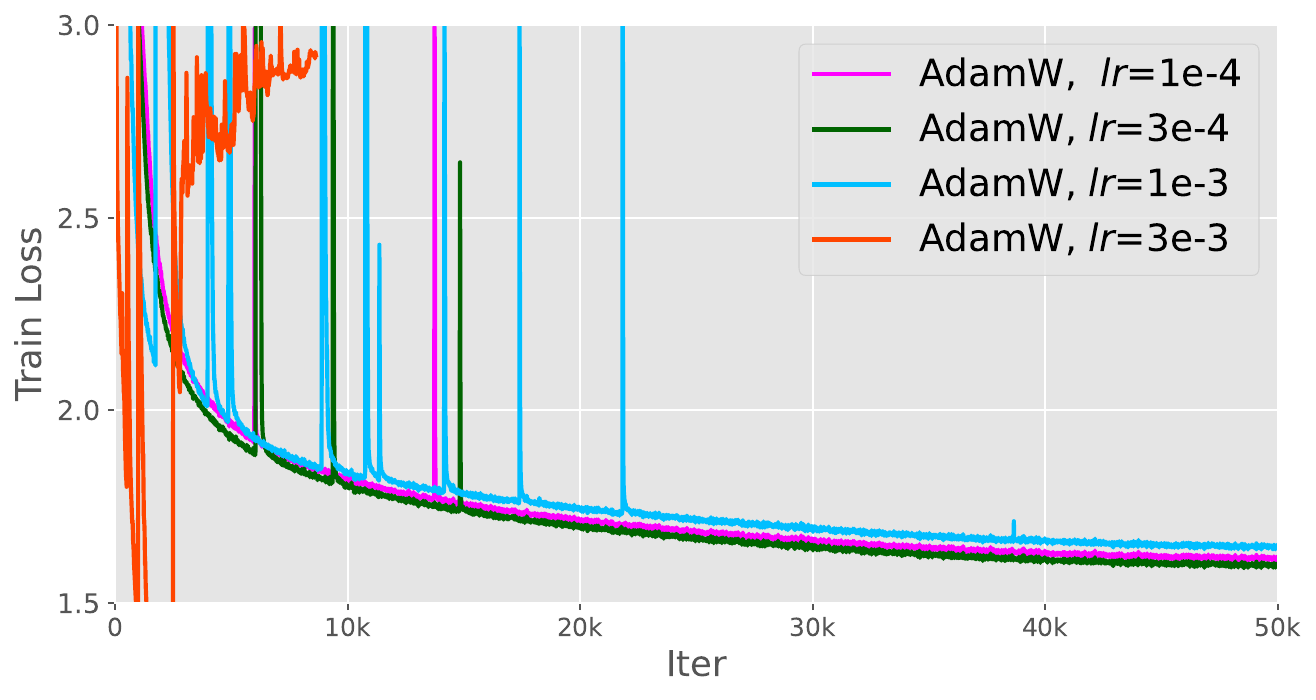}} \hspace{20 pt}
  \subfloat[\footnotesize{Adan} ]{\includegraphics[width=.42\textwidth]{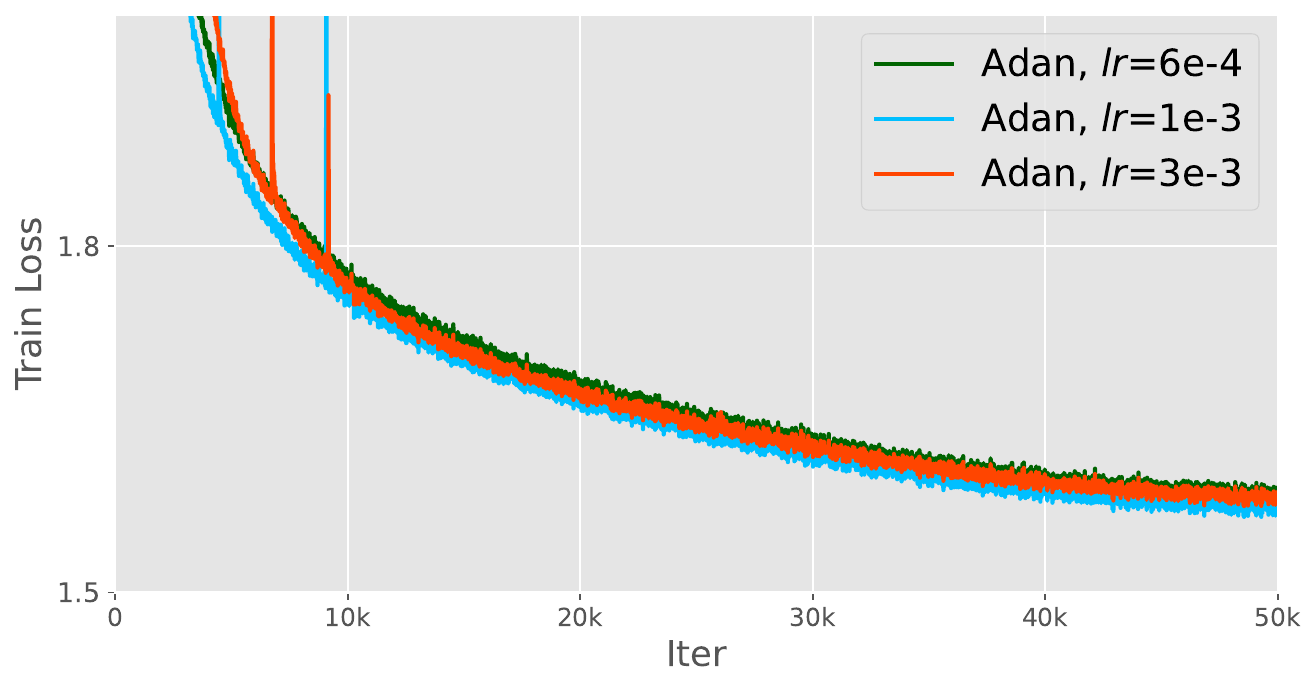}} \\
  \subfloat[\footnotesize{Lion}]{\includegraphics[width=.42\textwidth]{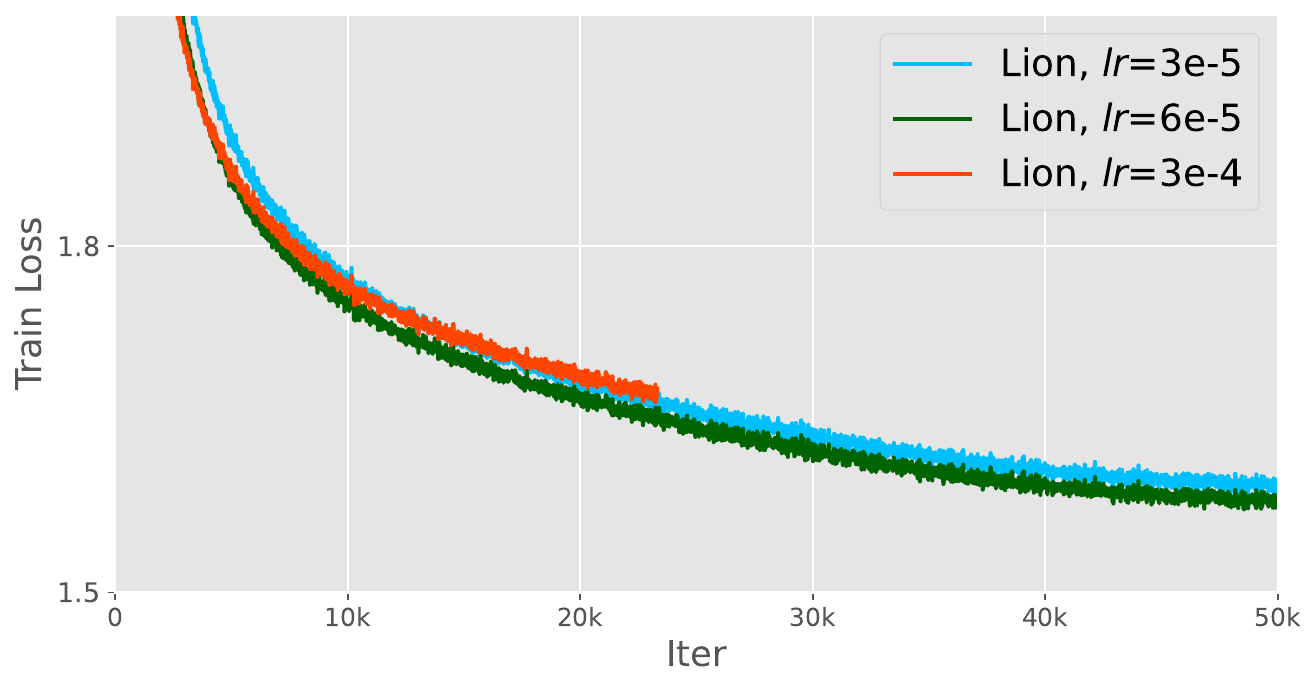}} \hspace{20 pt}
  \subfloat[\footnotesize{S3} ]{\includegraphics[width=.42\textwidth]{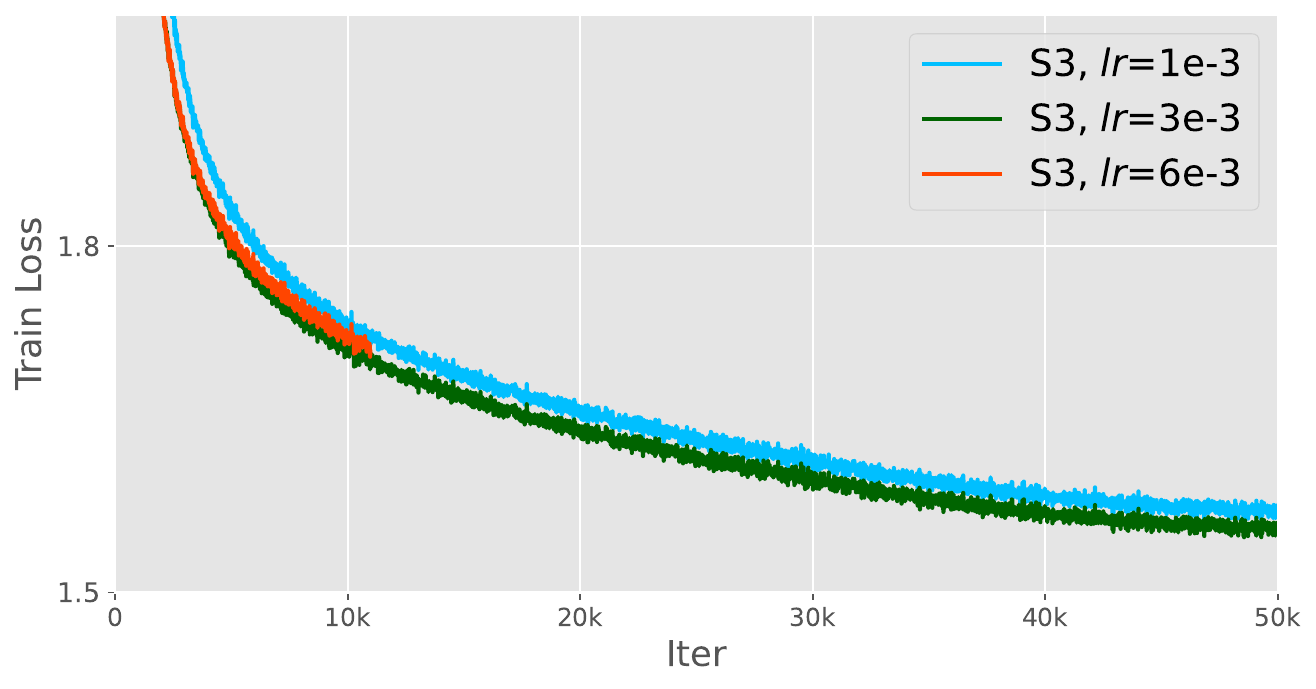}}
  \caption{\small{ Search the optimal peak learning  rates for \textsf{\footnotesize AdamW}, \textsf{\footnotesize Adan}, \textsf{\footnotesize Lion} and \textsf{\footnotesize S3} pre-training GPT-2 (345M)  on OpenWebText. } }
  \label{Fig.9}
\end{figure*}

We use the PyTorch vision reference codes \footnote{https://github.com/pytorch/vision/tree/main/references/classification} to implement vision tasks. For data augmentation, we adhere to the recommended settings in the codes, incorporating RepeatedAugment, AutoAugment Policy (magnitude=$9$), and Mixup($0.2$)/CutMix($1.0$) with a probability of $0.5$. Additionally, label-smoothing with a value of $0.11$ is applied. The batch size is set to $1024$ for ResNet-50 and $4096$ for ViT-B/16. Regarding the learning rate scheme, we linearly increase it to its peak in the initial $30$ epochs and then apply a cosine decay, decreasing it to $0$ in the subsequent epochs.
Other customized hyperparameters for \textsf{\small SGD} and \textsf{\small AdamW} are well-established in the codes,
and the settings for \textsf{\small Adan} and \textsf{\small lion} are followed the recommendations to train ResNet-50 and ViT-B/16 in their respective official papers \cite{Adan2024,Lion2023}. Since \textsf{\small NAdam} is similar to \textsf{\small AdamW}, so all its hyperparameters are also the same as \textsf{\small AdamW}.  Specifically, we list the hyperparameters of all the optimizers as follows:

\begin{itemize}
\item For \textsf{\small SGD}, we use $lr_{\max}=0.3, {\rm mom}=0.9$ , as is default value in the official codes.
  \item For \textsf{\small AdamW}, we utilize $lr_{\max}=0.003, \beta_1=0.9, \beta_2=0.999$ to train ResNet-50 and ViT-B/16, as is default value in the official codes and also widely used in other papers \cite{AdaBelief2020,Adan2024,Lion2023}.
  \item For \textsf{\small Adan}, we employe $lr_{\max}=0.015, \beta_1=0.98, \beta_2=0.92, \beta_3=0.99$, per official recommendations \cite{Adan2024}.
  \item For \textsf{\small Lion}, we adopt $lr_{\max}=0.001, \beta_1=0.9, \beta_2=0.99$, per official recommendations ite{Lion2023}.
  \item For \textsf{\small S3}, we set $lr_{\max}=0.006, \beta=0.95, p=3$. We conducted a coarse hyperparameters search on ViT-B/16  (Subsection \ref{Sensitivity Analysis})   and extended the hypermeters to ResNet-50 without further tuning.
\end{itemize}

 We utilize Megatron-LM \footnote{https://github.com/NVIDIA/Megatron-LM} to implement the language tasks. We use Megatron-LM to implement language tasks. Following the standard GPT-2 protocol, we construct a 345M Transformer decode-only model with the number of layers set to $12$, sequence length to $1024$, hidden size to $512$, and the number of attention heads to $8$. To testify the effectiveness of \textsf{\small S3} in training a productive LLM model, we additionally construct a large 7B model with the number of layers set to $32$, sequence length to $4096$, hidden size to $4096$, and the number of attention heads to $32$. GPT-2 (345M) is trained on OpenWebText with a batch size of $512$, and GPT-2 (7B) is trained on the refined CommonCrawl with a batch size of $1024$.  For \textsf{\small S3}, we set $p$ to $3$ and do not employ the gradient clipping technique. For other optimizers, the gradient clipping threshold is set to $1.0$.
Regarding the learning rate scheme, we linearly increase it to the peak in the initial $5k$ steps and then decrease to $0.1\times$ of the peak with a cosine decay in the subsequent steps.
The peak learning rates of all the optimizers are from coarse search for training GPT-2 (345M)(please refer to Figure \ref{Fig.9}). Other customized hyperparameters are listed below:

\begin{itemize}
  \item For \textsf{\small AdamW}, we utilize $\beta_1=0.9, \beta_2=0.95$, which are widely used in train LLMs \cite{GLM130B2022,Palm2023,Llama2023,Baichuan22023}.
  \cite{AdaBelief2020,Adan2024,Lion2023}).
  \item For \textsf{\small Adan}, we employe $\beta_1=0.98, \beta_2=0.92, \beta_3=0.99$, per official recommendations \cite{Adan2024} for train LLMs.
  \item For \textsf{\small Lion}, we adopt $\beta_1=0.95, \beta_2=0.98$, per official recommendations \cite{Lion2023} for train LLMs.
  \item For \textsf{\small S3}, we set $\beta=0.95, p=3$. We conducted a coarse hyperparameters search on ViT-B/16  (Subsection \ref{Sensitivity Analysis})   and extended the hypermeters to train LLMs without further tuning.
\end{itemize}

Notably, we followed the weight decay adjustment strategy outlined in the paper \cite{Lion2023}. Specifically, we used the product of the peak learning rate ($lr_{\rm Adam}$) and the weight decay ($\lambda_{\rm Adam}$) from \textsf{\small AdamW} as a constant. For other optimizers, we just determine the peak leaning rate, and the weight decay was derived directly using the formula $\lambda = \frac{lr_{\rm Adam}\lambda_{\rm Adam}}{lr}$. Importantly, the the baseline peak learning rates and weight decays of Adam  for training our ResNet-50 and ViT-B-16 are also the same with those reported in \cite{Lion2023}, while that for GPT-2  are the same with the paper on Llama \cite{Llama2023}.

\newpage

% your own custom macro containing the $\backslash${\tt{includegraphics}} command to make things
%\bf{If you include a photo:}\vspace{-33pt}
%\begin{IEEEbiography}[{\includegraphics[width=1in,height=1.25in,clip,keepaspectratio]{fig1}}]{Michael Shell}
%Use $\backslash${\tt{begin\{IEEEbiography\}}} and then for the 1st argument use $\backslash${\tt{includegraphics}} to declare and link the author photo.
%Use the author name as the 3rd argument followed by the biography text.
%\end{IEEEbiography}

\vfill

\end{document}